\documentclass[letterpaper, 10 pt, conference]{ieeeconf}  % Comment this line out if you need a4paper

\usepackage{graphicx} % Required for inserting images
\usepackage{cite}
\usepackage{float}
\usepackage{hyperref}
\usepackage{caption}
\usepackage{authblk}
\usepackage{rotating}
\usepackage{dblfloatfix}
\usepackage{pifont}% http://ctan.org/pkg/pifont

\title{CRASAR-U-DROIDs: A Large Scale Benchmark Dataset for Building Alignment and Damage Assessment in Georectified sUAS Imagery}

\author{Thomas Manzini}
\author{Priyankari Perali}
\author{Raisa Karnik}
\author{Robin Murphy}
\affil[1]{Department of Computer Science and Engineering, Texas A\&M University}
\affil[ ]{\textit {\{tmanzini, perali, raisak, robin.r.murphy\}@tamu.edu}}

\begin{document}

\maketitle

\begin{abstract}
   This document presents the Center for Robot Assisted Search And Rescue - Uncrewed Aerial Systems - Disaster Response Overhead Inspection Dataset (CRASAR-U-DROIDs) for building damage assessment and spatial alignment collected from small uncrewed aerial systems (sUAS) geospatial imagery. This dataset is motivated by the increasing use of sUAS in disaster response and the lack of previous work in utilizing high-resolution geospatial sUAS imagery for machine learning and computer vision models, the lack of alignment with operational use cases, and with hopes of enabling further investigations between sUAS and satellite imagery. The CRASAR-U-DRIODs dataset consists of fifty-two (52) orthomosaics from ten (10) federally declared disasters (Hurricane Ian, Hurricane Ida, Hurricane Harvey, Hurricane Idalia, Hurricane Laura, Hurricane Michael, Musset Bayou Fire, Mayfield Tornado, Kilauea Eruption, and Champlain Towers Collapse) spanning 67.98 square kilometers (26.245 square miles), containing 21,716 building polygons and damage labels, and 7,880 adjustment annotations. The imagery was tiled and presented in conjunction with overlaid building polygons to a pool of 130 annotators who provided human judgments of damage according to the Joint Damage Scale. These annotations were then reviewed via a two-stage review process in which building polygon damage labels were first reviewed individually and then again by committee. Additionally, the building polygons have been aligned spatially to precisely overlap with the imagery to enable more performant machine learning models to be trained. It appears that CRASAR-U-DRIODs is the largest labeled dataset of sUAS orthomosaic imagery. 
\end{abstract}

\section{Introduction}
\begin{figure}
\includegraphics[width=\columnwidth]{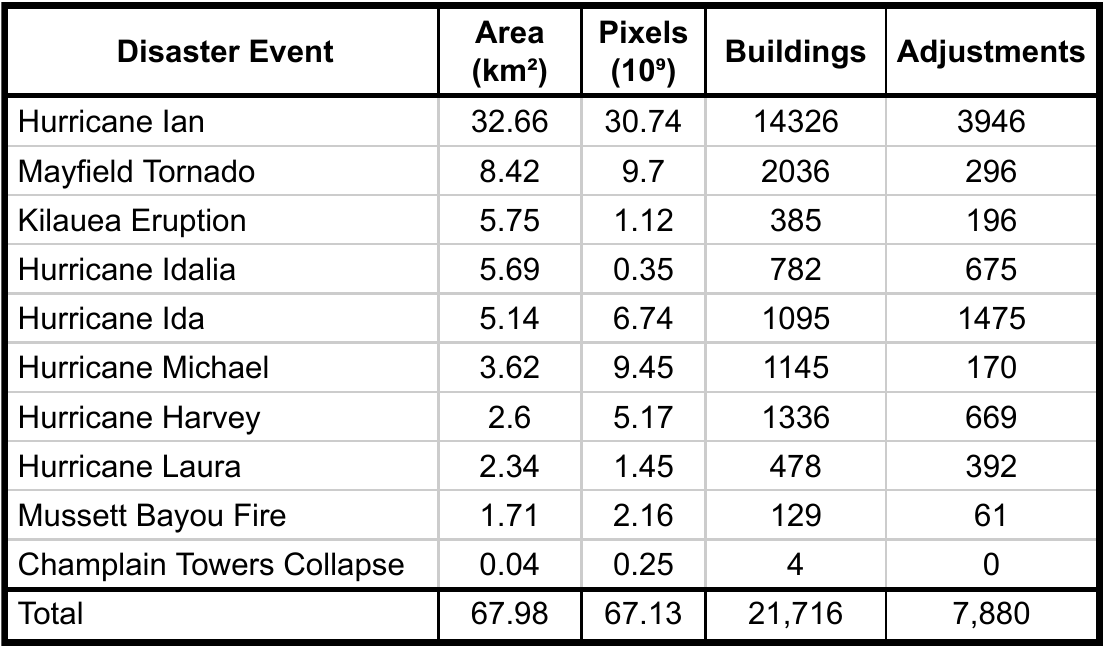}
\captionof{table}{The composition of the CRASAR-U-DROIDs dataset split by disaster events showing the total area in square kilometers, the total number of pixels in gigapixels, the total number of buildings, and the total number of adjustments. The disaster events are ordered in decreasing order of total area.\label{Tab:dataset_split_by_disaster}}
\end{figure}

%The \underline{C}enter for \underline{R}obot \underline{A}ssisted \underline{S}earch And \underline{R}escue - \underline{U}ncrewed Aerial Systems - \underline{D}isaster \underline{R}esponse \underline{O}verhead \underline{I}nspection \underline{D}ata\underline{s}et 
The Center for Robot-Assisted Search and Rescue - Uncrewed Aerial Systems - Disaster Response Overhead Inspection Dataset 
(CRASAR-U-DROIDs) is a dataset of georectified orthomosaic imagery collected during the response to federally declared disasters in the United States and supplemented by building polygons that have been spatially aligned to the imagery and labeled for their building damage category based on the Joint Damage Scale (JDS)\cite{gupta2019xbd}. This dataset covers 52 orthomosaics collected from 10 federally declared %collapse was NOT a wide-area disaster
disasters spanning 67.98 square kilometers (26.245 square miles), 
%why acres? use international units
containing 21,716 building polygons and damage labels, and 7,880 adjustment annotations. A breakdown of this data by disaster event is shown in Table \ref{Tab:dataset_split_by_disaster}. CRASAR-U-DROIDs represents the largest published dataset in the scientific literature of labeled orthomosaics collected from small uncrewed aerial systems (sUAS) in terms of pixels and building count. The dataset is publicly available at 
\url{https://huggingface.co/datasets/CRASAR/CRASAR-U-DROIDs}. 
The dataset was collected by CRASAR member researchers at either Florida State University or Texas
A\&M University while working directly for agencies having jurisdiction at the disaster in
some combination of roles as the UAS coordinator,
provider of drones and pilots, or as drone data managers and thus is representative of actual operations. All imagery was screened for any ethical violations, such as personal identifiable information, and the provenance is known, providing transparency and accountability. 

% FAILS THE YELLOW HIGHLIGHTER TEST Specifically, this dataset consists of imagery collected at Hurricanes Harvey, Idalia, Ian, Michael, Ida, and Laura, the Kilauea Volcano Eruption, the Mussett Bayou Fire, the Champlain Towers Collapse, and the Mayfield Tornado. This large, real-world dataset is expected to support research that will lead to machine learning models that could be adopted by emergency managers.

The motivation for this dataset is two-fold, with the ultimate goal to enable computer vision/machine learning (CV/ML) techniques to be effectively used during disaster operations by emergency managers. The primary motivation is to facilitate the use of CV/ML with sUAS imagery, as the ML community has typically worked with satellite imagery and the sUAS community is only beginning to explore ML. 
Low-cost sUAS are becoming an important tool for disaster response as they can be directly tasked by responders immediately after the event rather than through requesting strategic or costly resources from satellites or high-altitude crewed aircraft providers, both of which may take days to deploy and distribute imagery \cite{manzini2023quantitative, fernandes2018quantitative, fernandes2019quantitative}. Much work has been done collecting data from satellite imagery \cite{fujita2017damage, gupta2019xbd, haitiBRDDataset, IdaBDDataset}; however, the ML models trained on satellite data are either not able to handle or not able to effectively leverage the high-resolution imagery that is collected from sUAS systems.
The second motivation is to facilitate cross-evaluation of CV/ML models trained on disaster imagery from different sensing platforms and at different ground sample distances. This will enable the CV/ML community to benchmark maturing techniques for managing variable scale imagery.

%\textbf{while also providing a dataset of sUAS data with greater spatial coverage to the CV/ML communities? HUH?}. WHAT IS THE ADVANTAGE OF THIS FOR CV/ML? FOR DISASTER RESPONSE. 

While there exists other sUAS labeled imagery datasets for disasters, notably RescueNet\cite{rahnemoonfar2022rescuenet}, FloodNet\cite{rahnemoonfar2021floodnet}, Volan v.2018 \cite{Pi2020}, ISBDA\cite{zhu2021msnet}, and DoriaNet\cite{cheng2021dorianet}, those datasets collectively have four limitations that the CRASAR-U-DRIODs dataset overcomes:
\begin{itemize}
\item The small scale of the prior datasets in terms of coverage area, pixels, and other attributes. Small coverage areas mean fewer spatial features and information, and either fewer pixels or fewer unique pixels. Due to the high-resolution sUAS imagery provides, and the low altitudes at which they are typically flown, sUAS imagery naturally captures a much smaller area in comparison to manned or satellite imagery. ML modeling efforts may benefit from the additional information present in the high-resolution sUAS imagery but are disadvantaged by the lack of spatially diverse imagery which limits modeling efforts and eventual generalization. As will be described more in detail in section \ref{sec:motivation_prev_work}, the CRASAR-U-DRIODs dataset provides a larger scale of data in comparison to the prior work. 

\item The lack of diversity of disasters, in terms of types and number, which hinders the generalization of ML models to new disaster types and events due to a lack of representative data. As discussed in section \ref{sec:imagery_acquistion}, the CRASAR-U-DRIODs dataset provides imagery for ten federally declared disasters, consisting of five different types of disasters. This will benefit future ML modeling efforts in tasks like building damage assessment as they can be evaluated on a larger set of disaster events and types. This will result in ML models which will align more closely to operational needs in practice.

\item The inability to transfer ML techniques from satellite datasets to sUAS datasets due to varying label schemas for the same tasks. Without the use of the same label schema for building damage assessment, ML techniques used for satellite imagery cannot be evaluated on sUAS datasets without augmentation; however, as described in section \ref{sec:BDA}, the CRASAR-U-DRIODs dataset utilizes the same JDS annotation schema for building damage to allow for transfer of such techniques.

 \item The lack of spatial alignment of preexisting spatial data, such as building polygons and road lines, with sUAS imagery. Spatial alignment errors in sUAS imagery are non-uniform \cite{manzini2024non}; therefore, previous efforts to uniformly correct for misalignment, such as\cite{gupta2019xbd}, are not sufficient to handle the sUAS spatial misalignment case. CV/ML efforts to develop performant building damage assessment models utilizing a priori building polygons, such as the building polygons from Microsoft Building Footprints \cite{MicrosoftBuildingFootprints}, would be impeded by this misalignment. The building polygons in the CRASAR-U-DROIDs dataset have been manually aligned, as described in section \ref{sec:spatial_alignment}.
\end{itemize}

The remainder of this document is organized as follows.  Section~\ref{sec:motivation_prev_work} summarizes the previous work in sUAS, manned aircraft, and satellite imagery datasets for disasters. Section~\ref{sec:crasar-dataset} describes the dataset creation, including the imagery acquisition, building damage assessment annotation process, the spatial alignment data, and the dataset composition. The limitations of the dataset and annotation process are discussed in section~\ref{sec:limitations}, with section~\ref{sec:contributions} concluding with the contributions and current efforts with regards to the dataset.

\begin{figure} 
\centering
\includegraphics[width=0.49\textwidth]{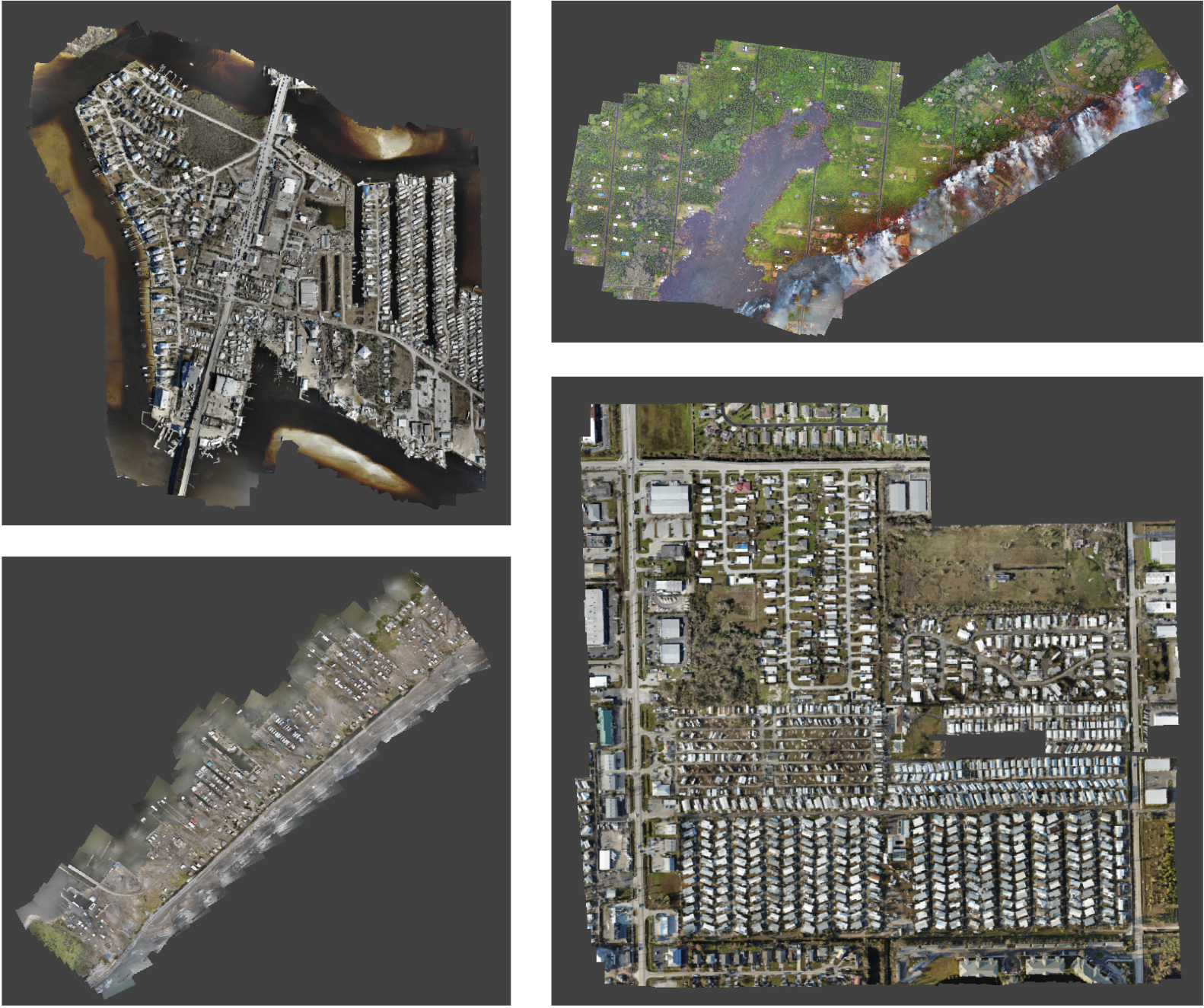}
\caption{Four sample orthomosaics from the fifty-two (52) orthomosaics within the CRASAR-U-DROIDs dataset.}
\label{Fig:ortho_sample}
\end{figure}

\section{Prior Work}
\label{sec:motivation_prev_work}

% USUALLY THE TOPIC PARAGRAPH IS 1 PARAGRAPH, MAYBE A COUPLE
% REPEAT: AVOID ONE SENTENCE PARAGRAPHS. THIS DOES NOT MATCH THE SUBSECTIONS!
%This section discusses relevant prior work along two thrusts: aerial imagery datasets for disaster response purposes and orthomosaics collected from sUAS. 

\begin{figure*}[h]
\centering
\includegraphics[width=0.98\textwidth]{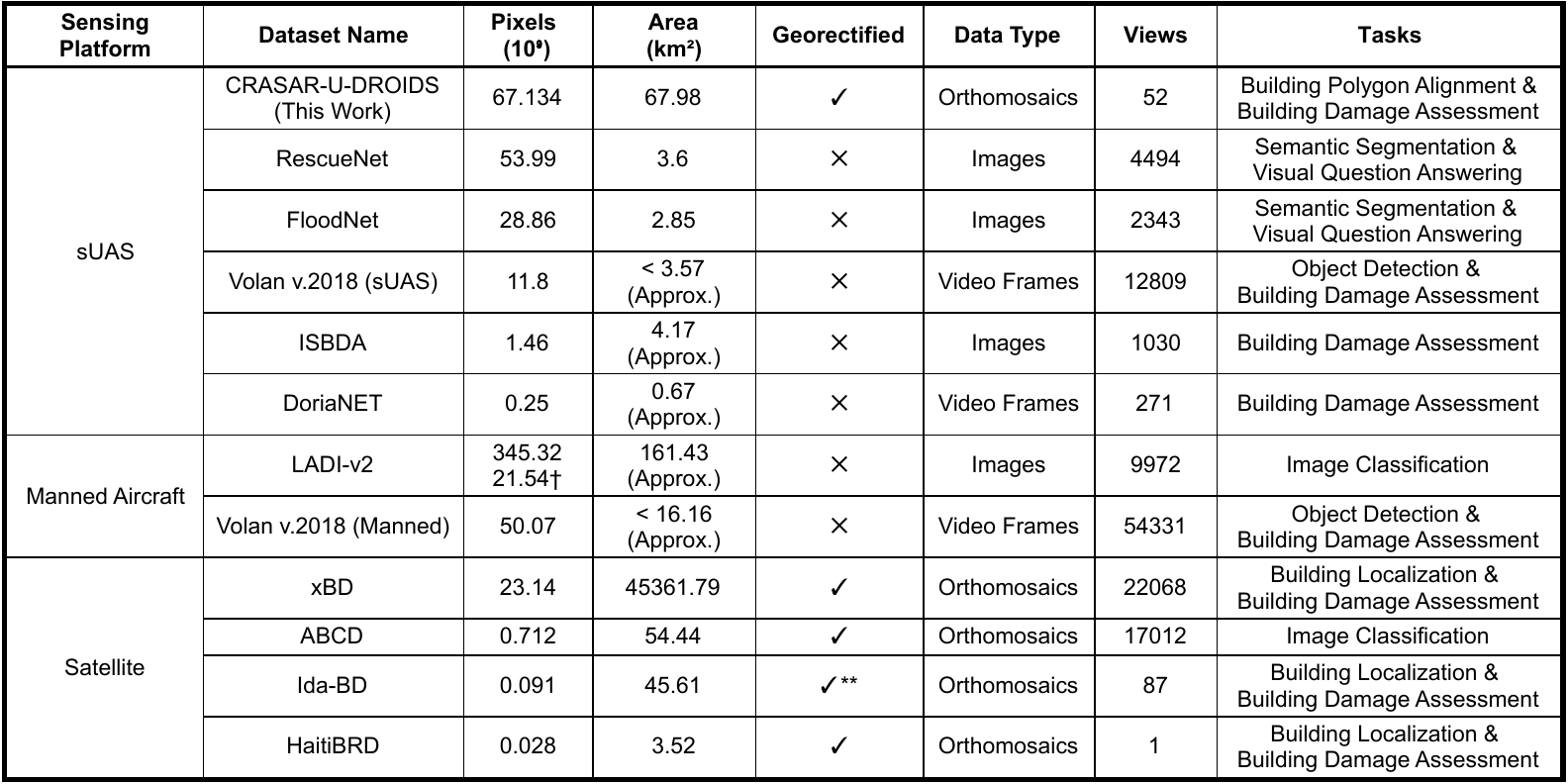}
\captionof{table}{Table comparing other aerial computer vision datasets collected from disasters to the CRASAR-U-DROIDs dataset. Datasets are ordered first by the sensing platform (sUAS, Manned Aircraft, Satellite), followed by pixel count to indicate approximate sizes. Within the subset of sUAS datasets, this work represents the largest dataset in terms of pixels and spatial area. Further, it is the only sUAS disaster dataset that is orthorectified. 
(\textdagger Denotes an author recommendation to utilize the smaller, resized, version of their dataset, ** Indicates that the released data has had the coordinate reference system transform removed.)\label{Tab:disaster_dataset_comparison}}
\end{figure*}

A review of literature identifies five sUAS datasets specifically from disasters, but have gaps which are addressed by CRASAR-U-DROIDs:
they are not georectified;
represent a much smaller scale of spatial area (67.98km\textsuperscript{2} vs approximately 4.17km\textsuperscript{2}),  pixels (67.13 gigapixels vs 53.99 gigapixels), represent a smaller number of disasters (10 vs 6), and fewer types of disasters (5 vs 2); %largest
have less  utility for emergency management because the labeling schemas were ad hoc or a misapplication of an existing standard;
and do not consider spatial alignment. 
%because they annotated raw imagery.
These five datasets are a subset of a total of twelve existing labeled disaster imagery datasets 
from any aerial asset (4 unmanned, 3 manned, 4 satellite, and 1 both manned and unmanned) identified 
via a systematic search within the IEEE Xplore, ACM Digital Library, Science Direct scientific databases, and the natural hazards DesignSafe Data Depot \cite{DesignSafe}. The criteria for choosing the 12 datasets was to include only those with i) aerial imagery from disasters with significant economic consequences which overwhelmed local response capabilities, consistent with the definition of a disaster, ii) labeled damage to populated areas and structures rather than attributes such as boundaries of affected areas, in keeping with the CRASAR-U-DROIDs focus on building damage assessment iii) the dataset was at least partially available for download. Seventeen datasets failed to meet the criteria for inclusion: eight fires \cite{ribeiro2023burned, shamsoshoara2021aerial, prabowo2022deep, xu2024sen2fire, cambrin2024cabuar, WANG2024123489, ColombdaBurnedArea, chen2022wildland}, four floods \cite{montello2022mmflood, bonafilia2020sen1floods11, rambour2020flood, Nugraha2023AerialIO}, one landslide \cite{xu2024cas}, one earthquake \cite{cambrin2024quakeset}, and three which included multiple types of events \cite{kyrkou2019deep, weber2022incidents1m, lee2020assessing}. 
Six datasets of labeled georectified data collected from sUAS that 
are not explicitly from disasters \cite{DroneDeployDataset, xu2024cas, chen20223D, barros2022multispectral, zhong2020whu, niu2022hsi} are included in this review as a separate discussion on characterizing work on labeled georectified data, as such data is a key element of CRASAR-U-DROIDs.

\subsection{Aerial Imagery Disaster Response Datasets}
\label{sec:aerial_imagery_disaster_response}

\begin{figure*}[]
\centering
\includegraphics[width=0.98\textwidth]{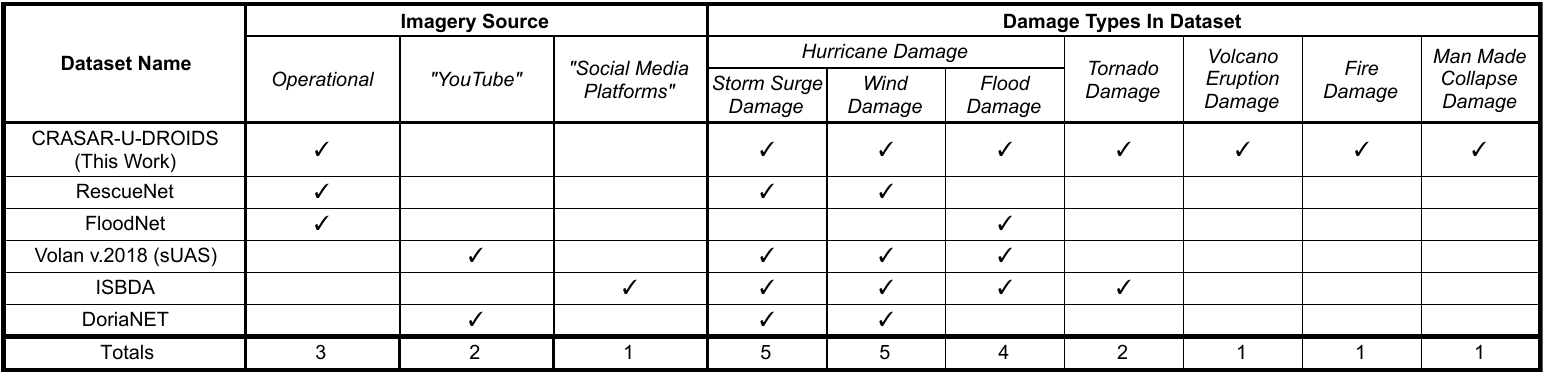}
\captionof{table}{
Comparison of other sUAS computer vision datasets collected from disasters to the CRASAR-U-DROIDs dataset based on the type of disaster damage present in each dataset.\label{Tab:dataset_disaster_types}}
\end{figure*}

\begin{figure}[!b]
\centering
\includegraphics[width=0.49\textwidth]{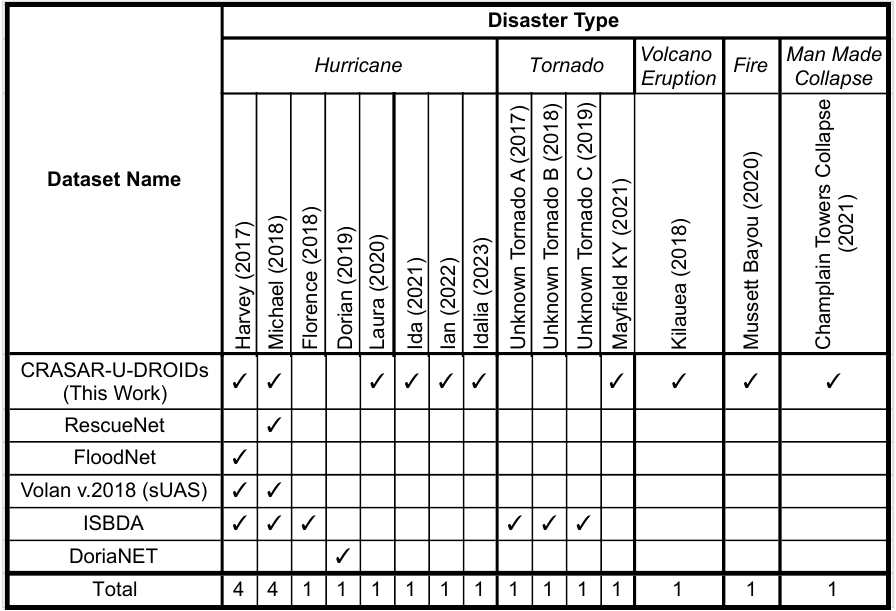}
\captionof{table}{
Comparison of the sources imagery from other sUAS computer vision datasets with the CRASAR-U-DROIDs dataset based on the specific disaster events represented.
\label{Tab:dataset_disaster_sources}}
\end{figure}

The 12 aerial imagery datasets for disaster response and assessment are divided into three categories based on the sensing platform that collected the imagery and detailed below: sUAS (Sec.~\ref{sec:suas_disaster_imagery_datasets}), manned aircraft (Sec.~\ref{sec:manned_aircraft_disaster_datasets}), and satellite imagery (Sec.~\ref{sec:satellite_datasets}). Each dataset is described in terms of the source event, the annotation schema, and the key attributes (area, tasks, etc.), which are summarized in Table~\ref{Tab:disaster_dataset_comparison}, with  Tables~\ref{Tab:dataset_disaster_types} and \ref{Tab:dataset_disaster_sources} providing additional details on the five sUAS disaster datasets. As Volan v.2018 \cite{Pi2020} contains imagery from multiple types of sensing platforms, it is discussed in multiple subsections. 

Table~\ref{Tab:disaster_dataset_comparison} is the overarching
comparison of 11 of the 12 datasets grouped by sensing platform and supports 
the claims that CRASAR-U-DROIDs is 
the only georectified sUAS disaster dataset and the largest  in terms of spatial area  and number of pixels. 
In order to facilitate the comparison, the key attributes (columns 2-8)  merit further explanation:
%and \ref{Tab:dataset_disaster_types}. 
%compares each of the datasets that will be discussed in this section.

%% NOTE: this table is the figure of merit for the entire paper, so it has to be explicitly explained 
%As Table \ref{Tab:disaster_dataset_comparison} summarizes the comparison of the 11 datasets with CRASAR-U-DROIDs, the key attributes (columns 2-8)  merit further explanation.
% I TRIED TO ADD THE "WHY DOES THIS IMPACT BDA" TO THIS, WHICH WA MISSING. 
% Tom: I changed this because BDA is only a portion of this dataset, with alignment being the other portion. Rewrote this to be more generic, focusing more on properties of ML systems rather than BDA models specifically.
\begin{itemize}
 %  \item Sensing Platform: Useful for understanding the potential altitudes, viewing angles, focal lengths, and sensors that could be used to generate imagery.
   \item \textit{Pixels}: Useful for comparing the rough scale of datasets, especially when imagery is georectified, as larger datasets can enable better model performance.
   
   \item \textit{Area}: Useful for comparing the spatial coverage of a dataset, as having many views of a small area may impede the generalization of downstream ML models.
   
    \item \textit{Georectified}: Useful for comparing the spatial uniqueness of pixels in the datasets, again providing a measure of dataset scale which may improve the generalization of downstream ML models.
     
    \item \textit{Data Type}: Useful for comparing how the imagery was collected and or processed.

    \item \textit{Views}: Useful for comparing the number of unique views of the different disaster scenes. The number refers to the count of images, video frames, or orthomosaics contained within a dataset. This number should be considered in conjunction with ``Data Type" and ``Pixels" to help determine the scale of the dataset.
  
   \item \textit{Tasks}: Useful for comparing the purpose of the dataset, the labels that accompany it, and how closely those labels support operational needs. 
\end{itemize}

\subsubsection{sUAS Disaster Imagery Datasets}
\label{sec:suas_disaster_imagery_datasets}

Each of the five sUAS disaster imagery datasets are recent (published between 2020-2022) and discussed in detail below, with 
the central points summarized in Tables \ref{Tab:dataset_disaster_types}, and \ref{Tab:dataset_disaster_sources}.
Table~\ref{Tab:dataset_disaster_types} lists the specific events from which the imagery for a dataset was taken and the source of that imagery.
While CRASAR-U-DROIDs contains imagery from Hurricanes Harvey and Michael, as does RescueNet, FloodNet, Volan v.2018, and ISBDA, it was either captured over different areas, at different altitudes, or by different drones than other datasets. The exceptions are FloodNet and RescueNet, which represent a subset of the raw imagery in CRASAR-U-DROIDs; however, this imagery has been georectified and annotated using a different schema.
Table~\ref{Tab:dataset_disaster_types} also highlights that researchers may not be paying sufficient attention to the quality and provenance of the imagery; it is hard to expect emergency managers to trust AI products based on YouTube videos supposedly taken by unknown sUAS at unnamed disasters. 
Table~\ref{Tab:dataset_disaster_sources} summarizes the specific name and category of the disaster event from which imagery was collected or attributed. It supports the claim that CRASAR-U-DROIDs covers the largest number of events (10) and types of events (5).

The FloodNet dataset \cite{rahnemoonfar2021floodnet} is a semantic segmentation and visual question-answering dataset of aerial images collected operationally by CRASAR in response to Hurricane Harvey. Notably, this dataset was used as a basis for a competition, both on semantic segmentation and visual question-answering, which garnered 102 submissions of ML models\cite{FloodNetResults}. Most relevant to this work were building-specific labels of ``building-flooded" and ``building-non-flooded." In total, this dataset contains 2,343 images, representing 28.86 gigapixels and an approximated 2.85km\textsuperscript{2} of area. This approximation was computed by generating orthomosaics from the imagery in the FloodNet dataset and computing the area of those orthomosaics.

The ISBDA dataset \cite{zhu2021msnet} is an object detection and segmentation dataset collected from ``social media" and contains imagery from sUAS flight over areas impacted by Hurricanes Harvey, Michael and Florence, as well as ``three tornados". 
%USED FOR? ANYBODY ELSE - Tom: There doesn't seem to be any work that has used this data in any meaningful way. Most of the citations use this work for context, specifically for the ML models they introduced.
While the authors claim that the dataset was annotated using the ``Joint Damage Scale", the damage levels of the buildings in these aerial images were annotated as either ``slight," ``severe," or ``debris" using both bounding boxes and masks. The ``debris" label corresponds to a fully collapsed building. These labels do not correspond to the Joint Damage Scale presented in \cite{gupta2019xbd}. In total, this dataset contains 1,030 images representing 1.46 gigapixels and approximately 4.17km\textsuperscript{2} of area. The area of this dataset was computed by assuming each sample represents one acre.

The Volan v.2018 effort \cite{Pi2020}, which was later extended in \cite{pi2021detection}, contains an object detection dataset collected from ``YouTube" and contains aerial imagery from overflights of areas impacted by Hurricanes Michael and Harvey. 
% WAS IT USED FOR ANYTHING? OTHER MODELS? Tom: Again, unfortunately, there doesn't seem to be any work that has used this data in any meaningful way. Beyond survey papers, and random contextual cites, the only work that used this data was from the PI's lab.
While this dataset contains imagery from both sUAS and helicopters, the sUAS portion will be discussed here. The imagery was annotated using bounding boxes, and the roofs of buildings were annotated as either ``Damaged Roof" or ``Undamaged Roof" with an additional category named ``debris". In total, the sUAS portion of this dataset contains 12,809 images representing 11.8 gigapixels and covering an approximated 3.57km\textsuperscript{2}. This approximation of area represents a reasonable upper bound as two of the four videos from the sUAS portion of this dataset are no longer available on YouTube. The approximation was arrived at by quadrupling the area of the CRASAR-U-DROIDs orthomosaic ``090403-Lancaster-Canyon-Gate.geo.tif" as the longest sUAS video in Volan v.2018, which remains available, was collected from the area within this orthomosaic.
%0.5mi\textsuperscript{2}

The DoriaNet Dataset \cite{cheng2021dorianet} is a segmentation dataset containing frames collected from sUAS video imagery posted to YouTube following Hurricane Dorian. 
% WAS IT USED FOR ANYTHING? OTHER MODELS? Tom: This was from the same lab as the Volan dataset, and again, unfortunately, there doesn't seem to be any work that has used this data in any meaningful way, the only work that used this data was from the PI's lab. 
The imagery was annotated using the FEMA HAZUS 2003 Hurricane Building Damage Scale \cite{hazusmh} using both building masks and bounding boxes. However, the authors use the Residential Building Damage Scale irrespective of building construction and purpose. As a result, it is unclear how this scale was applied practically, as the scale requires specific knowledge of the target building's construction and/or use. In total, this dataset contains 271 images representing 0.25 gigapixels and 0.67km\textsuperscript{2} of area. The area of this dataset was approximated by computing the area of a bounding polygon around the published sUAS flight path.

Finally, the RescueNet dataset \cite{rahnemoonfar2022rescuenet} is a semantic segmentation and visual question-answering dataset containing images collected by CRASAR during the response to Hurricane Michael. Notably, as with FloodNet, this dataset was used as a basis for a competition, which garnered 28 submissions of ML models\cite{RescueNetResults}. These images were annotated using 11 categories. Most relevant to this work are the ``building-no-damage", ``building-medium-damage", ``building-major-damage", and ``building-total-destruction" categories which were used to annotate buildings. In total, this dataset contains 4,494 images representing 53.99 gigapixels covering 3.6km\textsuperscript{2} of area. The area of this dataset was computed by generating an orthomosaic from the imagery in the RescueNet dataset and computing the area of that orthomosaic.

\subsubsection{Manned Aircraft Disaster Imagery Datasets}
\label{sec:manned_aircraft_disaster_datasets}

%WHAT IS THE TAKEAWAY?
%* different use cases, imagery
%* annotation format: volan bounding box, ladi- whole image  

%contextualizing- in what sense? 
%No outcomes are transferrable is to contrast with the fundamental differences Manned and Unmanned
%* higher altitude, spatial resolution (so wouldn't want to include in this dataset)
%* should be acknowledged
%* Volan appears both, LADI-v2 
%* larger area

%OBLIQUE, different distribution
%ladi many class classification- does this image contain a building with damage... different use case, screen images that meet category
%Note: our use case is fundamentally different and the information about the image (no GPS)

%volan: annotation of video... low and fast helicopter with chirons, varying focal lengths

Manned aircraft have been used for collecting imagery at disasters and three datasets have been identified; however these
datasets reflect quite different use cases (e.g., filtering images that show areas with damage), imagery properties (e.g., video not still images, oblique viewpoint not nadir), are not geo-tagged and so georectification would be impossible, and, most notably, employ different annotation
formats (esp. bounding boxes versus polygons or image-level labeling). As a result the datasets do not offer useful insights for constructing CRASAR-U-DROIDs, but are included for completeness. 

There are three relevant datasets for disaster response that were collected from manned aircraft. First, is the dataset presented in (Chen et. al. 2018) \cite{chen2018benchmark}, which annotated imagery from NOAA overflights of areas impacted by Hurricane Harvey using damage labels provided by a FEMA flood stage model. Second is the LADI dataset, which was originally released in 2019 but was recently updated (LADIv2) in 2024\cite{liuLargeScale2019}, provides data collected operationally by the Civil Air Patrol during ``federally declared disasters from 2009 onward" and was labeled according to the ``FEMA preliminary damage assessment criteria." Third, and finally, is a portion from the Volan v.2018 effort \cite{Pi2020}, which was collected from helicopters, obtained from ``YouTube", attributed to Hurricanes Irma, Maria, and Michael, and labeled using a schema defined by the authors. The rest of this section will further detail each of these datasets in the same order.

The (Chen et. al. 2018) dataset \cite{chen2018benchmark, choe2018benchmark} contains bounding box and building polygon annotations of building damage for areas impacted by Hurricane Harvey. 
The publication claims their annotations represent a combination of annotations from crowd workers and from FEMA flood damage model estimates as well as both satellite and manned aerial imagery\cite{chen2018benchmark}. However, only annotations for building damage from overflights of NOAA Manned Aircraft are available and referenced online \cite{choe2018benchmark}. Further, while annotations could be retrieved for 566,659 buildings, they do not appear to contain the crowd-sourced annotations that are described in the publication. Instead, only labels that correspond to the FEMA flood damage model estimates are available. All bounding boxes are annotated as either ``none", ``AFF," ``MIN," ``MAJ," or ``DES". As a result, it is difficult to accurately assess the size and relevance of this dataset as it is not fully available.
Due to this lack of certainty, this dataset has been omitted from Table \ref{Tab:disaster_dataset_comparison}.

The v2 version in the LADI dataset series \cite{liuLargeScale2019} is a 15-class multi-class image classification dataset containing high-resolution aerial images collected by occupants of manned aircraft performing overflights of ``federally declared disasters from 2009 onward" through The United States Civil Air Patrol. Of the 15 class labels that were applied to images, six corresponded to building labels, 4 of which corresponded to the FEMA preliminary damage assessment classes of ``affected", ``minor", ``major", and ``destroyed"; one which denoted flooded buildings; and one of which denoted the presence of buildings of any kind. In total, the v2 version of this dataset contained 9,972 images, representing 345.32 gigapixels and covering an approximated 161.43km\textsuperscript{2} of area. The area of the dataset was approximated by assuming each sample represents four acres. It should be noted that the authors recommend resizing the dataset to contain 21.54 gigapixels.

The portion of the Volan v.2018 effort \cite{Pi2020} contains imagery collected from manned aircraft over areas impacted by Hurricanes Irma, Maria, and Michael. In total, it contains 54,331 images representing 50.07 gigapixels and covering an approximated 16.16km\textsuperscript{2}. This approximation of area represents a reasonable upper bound as one of the four videos from the manned portion of this dataset is no longer available on YouTube. The approximation was arrived at by multiplying the area of the CRASAR-U-DROIDs orthomosaic ``10142018-MexicoBeach.geo.tif" by eight as it contains the area viewed in one of the remaining videos.

\subsubsection{Satellite Disaster Imagery Datasets}
\label{sec:satellite_datasets}

%* worth discussing because here are some models that could be plugged into our dataset because our dataset is georectified. as future work
%* shows symmetry, deliberate-- gave JDS 
%* xBD is used to train models, influential: future work- leverage all data
%<manned requires georectified>

Although manned aircraft disaster imagery datasets did not contribute to the development of CRASAR-U-DROIDs, one of the four identified satellite imagery datasets, xBD \cite{gupta2019xbd}, contributed the Joint Damage Scale used for annotations (see Sec.~\ref{subsec:annotation}) and models developed for the satellite datasets may be extensible to CRASAR-U-DROIDs since all are georectified and operate on orthomosaics (refer to Table~\ref{Tab:disaster_dataset_comparison}). 

%REORDER WITH XBD FIRST
Four efforts have also been made to develop datasets of disaster scenes collected by satellites. 
%TTHAT HELPS OUR WORK HOW? - Reworked to highlight that this enables the use of spatial data
These efforts are worth discussing specifically because all are georectified which enables the utilization of spatial data, like building polygons, in the same manner as CRASAR-U-DROIDs. First, is the benchmark dataset xBD \cite{gupta2019xbd}, which sourced imagery from MAXAR's Open Data Portal and was labeled using the Joint Damage Scale which was introduced by the authors; xBD covers 19 disasters detailed below. Second is the ABCD dataset \cite{fujita2017damage}, collected over areas impacted by the 2011 Japanese Tsunami sourced from the ``Pasco Image Archive" and labeled using a schema provided by the ``Japanese Ministry of Land, Infrastructure, Transport, and Tourism." Third is the Ida-BD dataset \cite{kaur2023large, IdaBDDataset}, which was sourced from MAXAR's Open Data Portal for Hurricane Ida and was labeled using an unknown schema. Finally, is the HaitiBRD dataset \cite{haitiBRDDataset}, sourced from MAXAR's Open Data Portal and labeled using an unknown label schema.
The rest of this section will further detail each of the datasets, by discussing the xBD dataset first, followed by the remaining datasets in chronological order of release.

The xBD dataset \cite{gupta2019xbd}, released in 2019, contains 22,068 orthorectified images collected over disaster scenes both before and after 19 disaster events. These 19 events contain 4 Hurricanes (Michael, Harvey, Florence, Matthew), 2 Volcano Eruptions (Guatemala, Lower Puna), 5 Wildfires (Santa Rosa, Woolsey, Pinery, Portugal, Carr), 2 Floods (Bangladesh, Midwest), 2 Tsunamis (Indonesia, Sunda Strait), 1 Earthquake (Mexico City), and 3 Tornados (Moore OK, Tuscaloosa AL, Joplin MO).
This dataset represents 23.14 gigapixels and spans 45361.79km\textsuperscript{2}. This dataset was focused on building damage specifically and utilized a five-class damage scale, termed the Joint Damage Scale (JDS), which contained the building damage labels ``no damage", ``minor damage," ``major damage," ``destroyed," and ``un-classified,"; the latter corresponding to buildings that are no longer present but were not believed to have been destroyed.

The ABCD dataset \cite{fujita2017damage}, released in 2017, is an image classification dataset of pre and post-disaster buildings, which were annotated for whether or not the building was ``washed away" by the 2011 Japanese Tsunami. Each building was annotated as either ``washed away" or ``surviving". In total, this dataset contains 17,012 orthorectified images representing 0.712 gigapixels covering 54.44km\textsuperscript{2}. 

Ida-BD \cite{kaur2023large, IdaBDDataset}, was released in 2022 following Hurricane Ida and contains 87 ortho rectified images spanning 45.61km\textsuperscript{2} and 0.091 gigapixels and was intended to be a novel test case for models trained on the xBD dataset \cite{kaur2023large, IdaBDDataset}. While the labels in the dataset and documentation correspond to the JDS, the damage scale is not explicitly stated.

HaitiBRD \cite{haitiBRDDataset}, was released in 2023 containing satellite imagery of the 2010 Hatian Earthquake and labeled buildings and roads for damage. For building damage, annotations again contained the damage labels that correspond to JDS (except for ``un-classified"), but it appears that there are no explicit statements of the damage scale utilized. In total, this dataset contained 1 orthorectified image representing 0.028 gigapixels, and 3.25km\textsuperscript{2}. 

\subsection{Labeled Georectified Data Collected From sUAS}
\label{sec:labeled_georectified_datasets}

%CONNECT THE DOTS
%* drone deploy is most widely cited, so what is different from ours?
%** smaller than ours 26 space and lower spatial resolution, not disasters, semantic segmentation 50 pixels
%* wide variety of use cases (ag, landslide), 
%* all are segmentation
%* what's different from us: all smaller except CAS, use cases (only disaster), we use building/road polygons  to constrain labeling
%* same: georectified, segmentation annotating pixels, 
%* also claim - largest for georectified sUAS not just for disasters CAS

Six labeled datasets of georectified sUAS imagery that did not cover disasters were found in the literature, and are typically smaller on both the area and pixel dimensions (refer to Table~\ref{Tab:disaster_dataset_comparison}), highlighting the unique size of CRASAR-U-DROIDs in general. The datasets cover a range of applications, specifically agriculture \cite{barros2022multispectral, zhong2020whu, niu2022hsi}, land use \cite{xu2024cas}, and urban scene semantic segmentation \cite{DroneDeployDataset, chen20223D} and do not use a priori maps, such as building or road polygons, to constrain labeling, thus avoiding
spatial alignment issues. 

%The spatial information captured in georectified data from sUAS represents an opportunity to build a bridge between sUAS imagery and satellite imagery. Such that, a system could leverage sUAS, satellite, and other georectified data sources jointly during training, or inference. However, there is little publicly available data from sUAS that possesses this spatial component and could realistically enable this use case. Each of the datasets within the literature will be further detailed within this section in order of relevance to the CRASAR-U-DRIODs dataset.
% WHERE IS THE RELATED WORK, THIS IS EXPOSITION.. AGAIN, FAILS THE YELLOW HIGHLIGHTER TEST --- modified paragraph above
%There are three labeled datasets of georectified sUAS data that were found in the literature, and below is a discussion of those datasets. 

The Drone Deploy Dataset \cite{DroneDeployDataset} provides orthomosaic imagery collected from sUAS over urban areas for non-disaster purposes. The Drone Deploy Dataset is a semantic segmentation dataset of 55 visual orthomosaics and elevation maps, which covered 2.43km\textsuperscript{2}, with a 10cm/px ground sample distance (GSD), resulting in 2.435 gigapixels of imagery \cite{DroneDeployDataset}. The CRASAR-U-DROIDs dataset differs by existing in the disaster response application area and at 
a higher resolution GSD while covering an order of magnitude more area.

The CAS Landslide Dataset \cite{xu2024cas} is a segmentation dataset of satellite and UAS orthomosaic imagery where areas of land impacted by landslides are annotated at the pixel level. This dataset contains 16 orthomosaics, 7 of which were collected by UAS. In total, this dataset contains 1.37 gigapixels of orthomosaic imagery collected over 4772.05km\textsuperscript{2}. The UAS portion of this dataset contains 0.883 gigapixels of orthomosaic imagery collected over 155.14km\textsuperscript{2} at a ground sample distance between 0.2 and 1 meter/pixel, a spatial resolution only slightly above the capabilities of current very high-resolution satellite imagery\cite{maxar15}. While this dataset captures landslides, a phenomenon that can result in disasters, this dataset focuses on detecting the landslide phenomenon rather than assessing its impact on populated areas. As a result, it is not considered a disaster dataset for this review of the literature.

The InstanceBuilding \cite{chen20223D} dataset is a collection of four annotated 3D georectified meshes constructed from imagery collected by sUAS. This data is also accompanied by annotated raw sUAS visual imagery. The georectified component of this dataset covers 0.44km\textsuperscript{2} and was generated using 3.562 gigapixels of imagery. Note that the pixels that are referenced here overlap between images, in contrast to other datasets in this section which pixels are all spatially unique.

Finally, three segmentation datasets containing multispectral orthomosaics collected from sUAS were found in the precision agriculture literature \cite{barros2022multispectral, zhong2020whu, niu2022hsi}. The dataset released in \cite{barros2022multispectral} contained 1.4 gigapixels of multispectral imagery covering 0.13km\textsuperscript{2} of area. The dataset released in \cite{zhong2020whu} contained 0.0145 gigapixels of multispectral imagery collected over 0.05km\textsuperscript{2} of area. Finally, the dataset released in \cite{niu2022hsi} collected 0.0036 gigapixels of multispectral imagery collected over 0.036km\textsuperscript{2} of area. All focus on the segmentation of crops in multispectral nadir imagery.

\subsection{Comparison Of CRASAR-U-DROIDs To Prior Work}

To summarize the analysis of previous work, 
the CRASAR-U-DRIODs dataset has four attributes (size, operational validity, accepted annotation schema, and adjustment for spatial misalignment)  that fill the gaps presented within the literature and distinguish it from other sUAS datasets.  In addition, the provenance of all imagery and post-processing for CRASAR-U-DROIDs is known, thus providing further transparency. 

First, this dataset is the largest labeled dataset of orthorectified sUAS imagery collected at disaster scenes. All prior work that released labeled datasets from sUAS at disaster scenes all focused on non-georectified, raw imagery \cite{kyrkou2019deep, rahnemoonfar2021floodnet, rahnemoonfar2022rescuenet, zhu2021msnet, cheng2021dorianet}. Though there has been work to leverage point clouds or other photogrammetry data products to perform building damage assessment following a disaster\cite{kerle2019uav}, no labeled datasets have been released that could be leveraged by others. One dataset of note is the dataset released in \cite{xu2024cas} which focuses on landslide area segmentation. While this dataset does contain more spatial area collected from sUAS than CRASAR-U-DROIDs (155.14km\textsuperscript{2} vs 67.98km\textsuperscript{2}), it is not in the same category. This is because the majority of landslides contained in \cite{xu2024cas} did not threaten populated areas, and the dataset focused on identifying landslide extent rather than assessing landslide impact on populated areas. In constrast, the CRASAR-U-DRIODs dataset contains imagery of populated areas impacted by disasters. 

% First, it is the first large-scale labeled dataset of orthorectified sUAS imagery collected at disaster scenes. While there has been work in building damage assessment from sUAS imagery, it has largely focused on non-georectified imagery \cite{kyrkou2019deep, rahnemoonfar2021floodnet, rahnemoonfar2022rescuenet, zhu2021msnet, cheng2021dorianet} or focused on point clouds or other photogrammetry data products at a smaller data scale or unsupervised methods on unlabeled data\cite{kerle2019uav}.

Second, these orthomosaics represent data collected and utilized for decision-making during the respective disaster response. The locations where these orthomosaics were captured were selected by the command element of the disaster during the response and recovery time frame.
This differentiates this dataset from others, which were captured opportunistically \cite{gupta2019xbd, IdaBDDataset, hansch2022spacenet} or collected from publicly available online content \cite{kyrkou2019deep, zhu2021msnet, cheng2021dorianet, Pi2020}, so these datasets do not necessarily capture the essential data used for decision-making. While \cite{rahnemoonfar2021floodnet} and \cite{rahnemoonfar2022rescuenet} also have this property, CRASAR-U-DROIDs captures this phenomenon across multiple disasters and geographic areas and has been labeled using a schema which aligns more closely with expected operational use cases.

Third, this dataset uses the JDS\cite{gupta2019xbd} to annotate building damage, allowing for the transfer of ML techniques between satellite and sUAS. The other datasets within both sUAS and manned do not use JDS, nor is there any other overlap in building damage label scales across datasets. With JDS being the same damage label scale utilized in the xBD dataset\cite{gupta2019xbd} and in this dataset, these two datasets are in the same label space, and the door is opened to training and testing on different spatial resolutions and on different sensing platforms (sUAS vs Satellite) at a scale not previously explored \cite{duarte2018satellite}. This has not been possible because of a fundamental lack of data that this dataset now provides.

Finally, it is the first dataset to explicitly address the spatial alignment errors that occur between imagery and spatial data. The tasks that the other datasets include do not address spatial alignment of any sort. There have been other datasets that have observed spatial alignment errors and have addressed them through uniform ``shifts" \cite{gupta2019xbd}; however, this dataset uniquely captures the non-uniformity of spatial alignment errors and provides another task of spatial alignment for buildings (building alignment) that has not been addressed previously. 

\section{The CRASAR-U-DROIDs Dataset}
\label{sec:crasar-dataset}

The CRASAR-U-DRIODs dataset's creation consisted of the acquisition of sUAS raw imagery from ten federally declared disasters (Sec.~\ref{sec:imagery_acquistion}) and generation of the 52 orthomosaics that make up the dataset (Sec.~\ref{subsec:rawimagery_ortho}), building damage assessment on acquired imagery (Sec.~\ref{sec:BDA}), and the corrections for spatial alignment errors observed (Sec.~\ref{sec:spatial_alignment}). 
%This section discusses the imagery acquisition at the ten wide-area disasters, the building damage assessment annotation process, the spatial alignment error correction process, and an overview of the dataset's composition in that order.  

% This section discusses the provenance of the imagery that composes the CRASAR-U-DROIDs dataset, followed by a discussion of all annotations and how they were reviewed. This section concludes with a discussion of the unique properties of this dataset compared with other work in the literature.

\subsection{Imagery Acquisition}
\label{sec:imagery_acquistion}

The imagery within this dataset was acquired through CRASAR's deployments at ten  disasters and converted to orthomosaics to align with the operational use and to lessen the format restrictions of such data. The CRASAR-U-DROIDs dataset does not represent all imagery collected during disaster operations, only select imagery from which orthomosaics could be constructed. This section first discusses the raw imagery acquisition of data at the ten federally declared disasters, followed by a discussion on the raw imagery conversion to the orthomosaics for the dataset. 

% The imagery present in this dataset was acquired through CRASAR's deployments to wide-area disasters. All imagery in this dataset was captured at the direction of the command elements of the disaster response. This does not represent all imagery collected during disaster operations, only imagery from which orthomosaics could be constructed.

\subsubsection{Disasters}

\begin{figure*}[]
\centering
\includegraphics[width=0.98\textwidth]{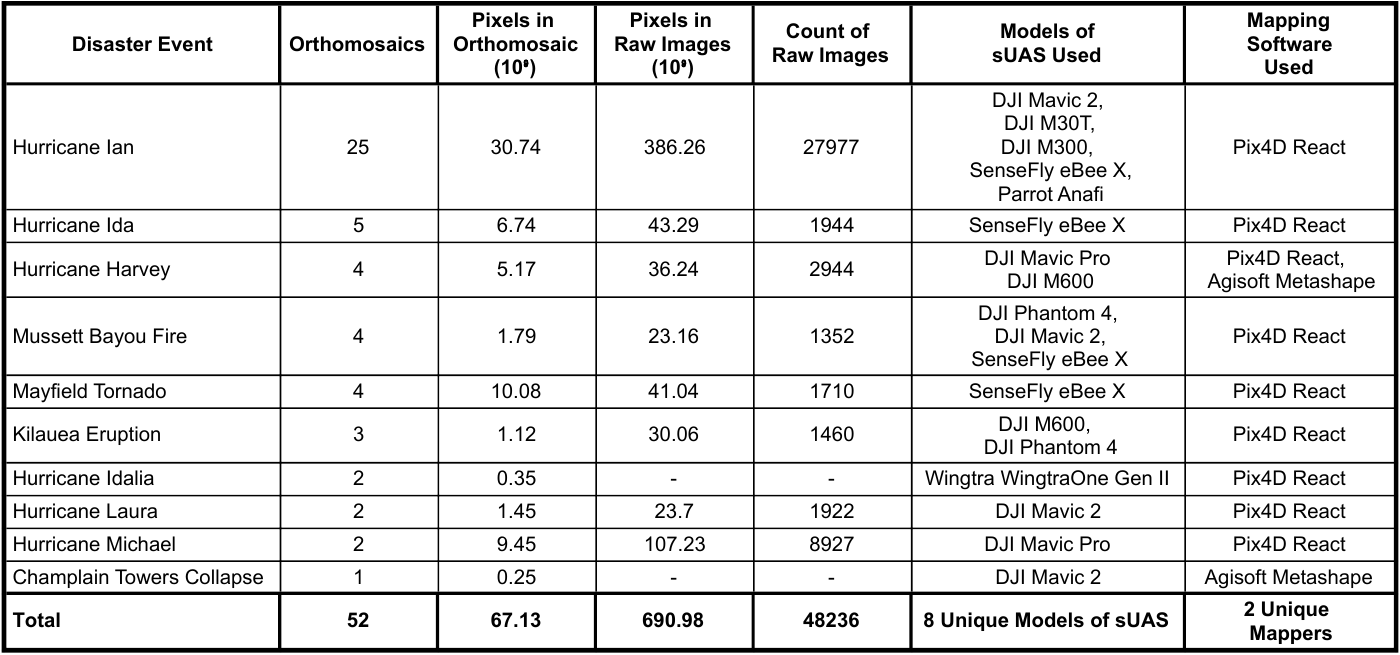}
\captionof{table}{This table characterizes the imagery present in the CRASAR-U-DROIDs dataset grouped by disaster at which it was collected and sorted by count of orthomosaics present in the dataset. The imagery collected at each disaster is characterized by the number of pixels in the associated orthomosaics, the number of pixels in and the count of raw images which were used to generate the associated orthomosaics, the models of sUAS which were used to collect the raw imagery, and the mapping software which was used to generate the associated orthomosaics. A dashline, ``-", indicates no data available for that column.\label{Tab:Disaster_imagery_Stats}}
\end{figure*}

The imagery within this dataset was captured at ten federally declared disasters, consisting of six hurricanes, one tornado, one volcano eruption, one wildland fire, and one building collapse through CRASAR's deployments. All imagery in this dataset was captured at the direction of the command elements of the disaster response. This does not represent all imagery collected during disaster operations, only imagery from which orthomosaics could be constructed. This section further describes each disaster and the raw imagery acquisition, with the descriptions provided in chronological order of disaster event's date. In total, at least 690.98 gigapixels of imagery, sourced from at least 48,236 images were used to generate the orthomosaics contained within the CRASAR-U-DROIDs dataset. The details of this discussion are shown in Table \ref{Tab:Disaster_imagery_Stats} for direct comparison.

Hurricane Harvey was a category 4 hurricane that made landfall in Texas, United States in August, 2017 \cite{HurricaneHarveyaNWS}. The orthomosaics associated with Hurricane Harvey were generated from 36.24 gigapixels (2,944 source images) of raw imagery collected with 2 different sUAS models, DJI Mavic Pro and DJI M600 between September 3, 2017 and September 4, 2017.  The sUAS operations at this event were documented in \cite{fernandes2018quantitative}.

The Kilauea Eruption was the volcanic eruption of the Kilauea Volcano in Hawaii, United States which started in April of 2018 \cite{neal20192018}. The orthomosaics associated with this eruption are generated from 30.06 gigapixels (1,460 source images) of raw imagery collected with 2 different models of sUAS, DJI M600 and DJI Phantom 4, on May 18, 2018. 

Hurricane Michael was a category 5 hurricane that made landfall in Florida, United States, in October 2018 \cite{HurricaneMicahelNWS}. The orthomosaics associated with Hurricane Michael were generated from 107.23 gigapixels (8,927 source images) of raw imagery collected via DJI Mavic Pro between October 13, 2018 and October 14, 2018. The sUAS operations at this event were further documented in \cite{fernandes2019quantitative}.

The Mussett Bayou Fire was a wildland fire in Walton County, Florida, United States, in May 2020 \cite{MussettNews}. The orthomosaics associated with the Musset Bayou Fire were generated from 23.16 gigapixels (1,352 source images) of raw imagery collected from 3 different types of sUAS models, DJI Phantom 4, SenseFly eBee X, and DJI Mavic 2, on May 8, 2020.

Hurricane Laura was a category 4 hurricane that made landfall in southeastern Texas and Louisiana, United States, in August 2020 \cite{HurricaneLauraNWS}. The orthomosaics associated with Hurricane Laura were generated from 23.7 gigapixels (1,922 source images) captured by DJI Mavic 2 on August 27, 2020.

The Champlain Towers Collapse was a multi-story residential collapse that occurred June 24, 2021 in Surfside, Florida, United States \cite{ChamplainTowersNIST}. One orthomosaic from this event was annotated for use in this dataset. This orthomosaic was collected on July 3, 2021 using a DJI Mavic 2. This orthomosaic was generated from an unknown number of source images because the source imagery could not be identified.

Hurricane Ida was a category 4 hurricane that made landfall in Louisiana, United States in August, 2021 \cite{HurricaneIdaNWS}. 43.29 gigapixels (1,944 source images) of raw imagery was used to generate the orthomosaics associated with Hurricane Ida. This raw imagery was collected between August 31, 2021 and September 2, 2021 by the SenseFly eBee X.

The Mayfield Tornado outbreak was a wide-area, federally declared disaster in which three tornadoes touched down in rapid succession in western Kentucky, United States, on December 10, 2021\cite{MayfieldTornadoNWS}. 
Three orthomosaics were collected at this event between the dates of December 13, 2021, and December 15, 2021, representing 41.04 gigapixels (1,710 source images) of raw imagery. It should be noted that this number excludes the raw imagery statistics that were used to generate the ``20211215-Russelville-Middle.geo.tif" orthomosaic for which the source imagery could not be identified. The raw imagery was collected with the  SenseFly eBee X.

Hurricane Ian was a category 4 hurricane which was later upgraded to category 5 which made landfall in Florida in September 2022\cite{HurricaneIanNWS}. The orthomosaics associated with Hurricane Ian were generated from 386.26 gigapixels (27,977 source images) of raw imagery which was captured using 5 different sUAS models: DJI M300, DJI M30T, SenseFly eBee X, DJI Mavic 2, and Parrot Anafi. The raw imagery was captured between the dates October 1, 2022 and October 2, 2022. At the time of writing, this deployment of sUAS represents the largest use of sUAS in a disaster to date \cite{manzini2023quantitative}.

Hurricane Idalia was a category 3 hurricane that made landfall in Florida, United States in August, 2023 \cite{HurricaneIdaliaNWS}. Two orthomosaics associated with Hurricane Idalia are included in this work. At the time of writing, an unknown amount of raw imagery was used to produce these orthomosaics as the raw imagery cannot be identified. However, it is known that the raw imagery was collected between August 30, 2023 and August 31, 2023 using a Wingtra WingtraOne Gen II.

% The imagery present in this dataset was collected by 9 different models of sUAS. In order of descending prevalence, these were the DJI Mavic 2 (17 orthomosaics), SenseFly eBee X (12 orthomosaics), DJI Matrice 30 (6 orthomosaics), DJI Mavic Pro (5 orthomosaics), DJI Phantom 4 (4 orthomosaics), DJI Matrice 600 (3 orthomosaics), DJI Matrice 300 (3 orthomosaics), Wingtra WingtraOne Gen II (2 orthomosaics), Parrot Anafi (1 orthomosaics). It should be noted that this differs only slightly from \cite{manzini2024non} in that a DJI Mavic 2 was used to capture a single orthomosaic that was not considered in \cite{manzini2024non}.

\subsubsection{Raw Imagery to Orthomosaics}
\label{subsec:rawimagery_ortho}

The raw imagery captured at the ten federally declared disasters via the sUAS models described earlier was converted to orthomosaics, resulting in 52 orthomosaics generated, samples of which are shown in Figure \ref{Fig:ortho_sample}. An intentional decision regarding the mapping software so as to align with the motivations and uses of this dataset. This section further discusses the reasoning behind converting the raw imagery to orthomosaics, the choice of mapping software, and a detailing of the resulting ground sample distances (GSDs) within the generated orthomosaics.  

\begin{figure*}
\includegraphics[width=\textwidth]{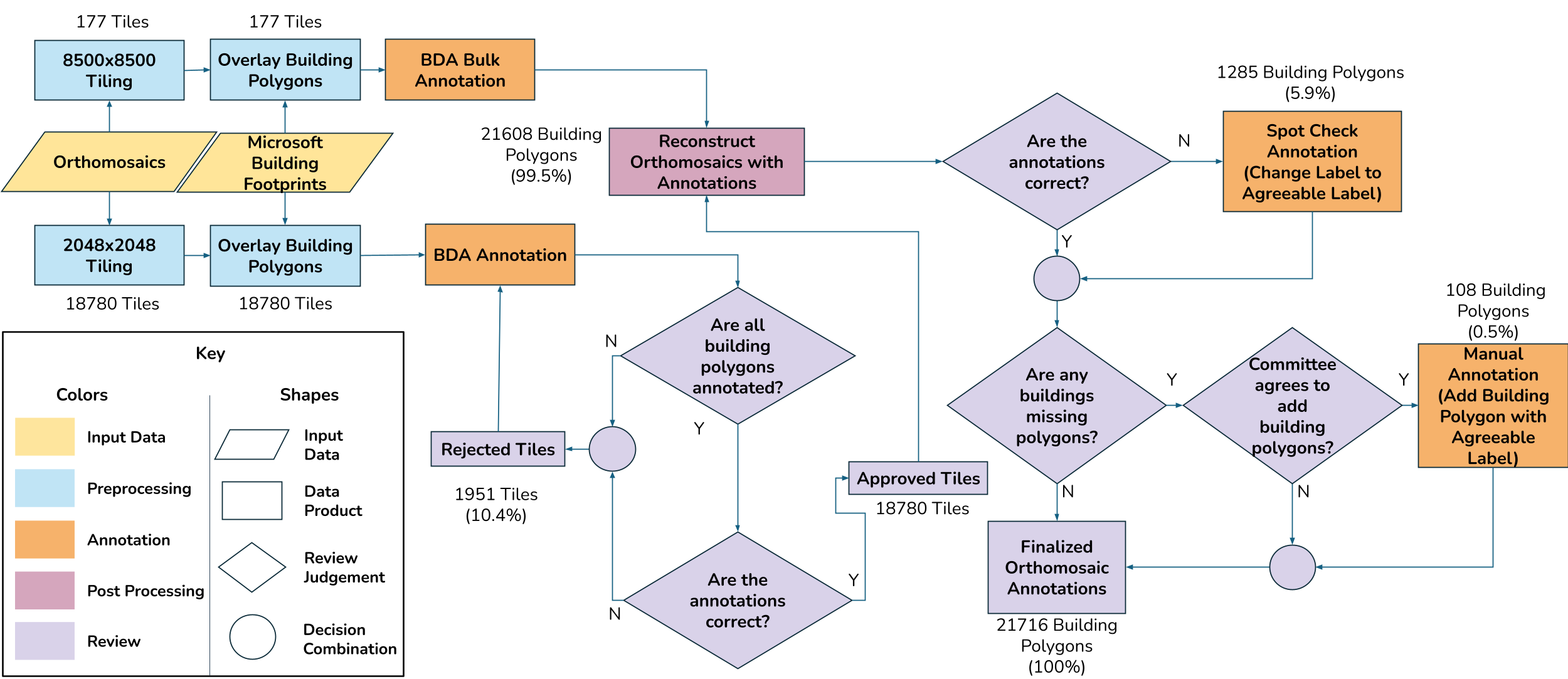}
\caption{The annotation and review workflow used to arrive at the joint damage scale (JDS) labels for the building polygons released here. For each building polygon, a single label was generated, which was then reviewed twice, once by a single reviewer and then a second time by a committee of the reviewers.}
\label{Fig:workflow}
\end{figure*}

The choice to provide orthomosaics, instead of image tiles, with this dataset was an intentional one. While previous datasets have provided images tiles or chips, as done with past satellite imagery \cite{gupta2019xbd, IdaBDDataset} instead of orthomosaics, the decision to include the complete orthomosaic is intended to not constrain users to predefined tile sizes and to enable data augmentation strategies utilizing variable dimension inputs which would be impeded by predefined image sizes. 

Following the capture of the raw data via the sUAS models described earlier, the imagery was converted to a georectified orthomosaic via mapping software, Pix4D React\cite{Pix4D} and Agisoft Metashape\cite{Agisoft}, and the choice of mapping software was intentional.
% \footnote{The raw data used to generate these orthomosaics has been retained but not hosted publically due to data scale. If you are interested in the raw data that was used to generate these orthomosaics, please contact the authors.}. 
Pix4D React was used as the mapping software of choice for two reasons. First, Pix4D React is specifically intended for use on edge computing devices, which could reasonably be expected to be deployed to the field alongside sUAS systems. Therefore, orthomosaic imagery generated via Pix4D React represents a variant of orthomosaic imagery that models trained on this data will likely encounter in practice.
Second, Pix4D React is specifically designed for rapid generation of orthomosaics at the expense of quality and accuracy, thereby representing the hardest variant of orthomosaic imagery that CV/ML systems could reasonably encounter. Within this dataset Pix4D React was used for 50 orthomosaics (96\%) and Agisoft Metashape for 2 orthomosaics (4\%). The reason that two orthomosaics were generated using Agisoft Metashape was because the raw data from which these orthomosaics were generated was not available at the time of annotation. As a result, orthomosaics from Pix4D React could not be obtained.

% Note that this dataset provides complete orthomosaics instead of image tiles or chips as have been done with satellite imagery in the past \cite{gupta2019xbd, IdaBDDataset}. Again, this was an intentional choice not to constrain users to predefined tile sizes and to enable data augmentation strategies utilizing variable dimension inputs which would be impeded by predefined image sizes.

% Following the capture of the raw data via the sUAS models described earlier, the imagery was converted to a georectified orthomosaic via mapping software, Pix4D React\cite{Pix4D} and Agisoft Metashape\cite{Agisoft}. Pix4D React was used for 50 orthomosaics (98\%) and Agisoft Metashape for 2 orthomosaics (4\%). 

Both mapping softwares, Pix4D React and Agisoft Metashape, generate orthomosaics at varying GSDs based on the resolution of the raw imagery. In the case of the CRASAR-U-DROIDs dataset, the generated 52 orthomosaics all vary in GSDs between 1.77 cm/px and 12.7 cm/px, with a mean of 3.74 cm/px. Further detailing of the GSDs by orthomosaics within the dataset is provided in Appendix Table \ref{Tab:Dataset overview}.

\subsection{Building Damage Assessment Annotation Process}
\label{sec:BDA}
The labeling process for the building damage assessment (BDA) element of the CRASAR-U-DROIDs dataset consisted of five primary steps: input data, preprocessing, annotation, post processing, and review. A visual overview of this workflow is shown in Figure \ref{Fig:workflow}. It should be noted that, throughout this workflow, the JDS \cite{gupta2019xbd} annotation schema, which consists of five damage labels: ``no damage", ``minor damage", ``major damage", ``destroyed", and ``un-classified", was used to align with the motivation of this dataset to allow for cross-evaluation of techniques between different sensing platforms and align with the operational use cases. This section further discusses the five steps within the workflow in the same order presented. 
%TODO ADD A DISCUSSION OF WHY JDS WAS CHOSEN AS THE DAMAGE SCALE OF CHOICE - Done
%A visual overview of the BDA workflow is shown in Figure \ref{Fig:workflow}.

\subsubsection{Input Data}
The BDA workflow took two sets of data as input, the orthomosaics and Microsoft Building Footprints. The orthomosaics were sourced from the orthomosaics that were generated from the raw imagery, discussed in section \ref{subsec:rawimagery_ortho}. The Microsoft Building Footprints were sourced from the Microsoft Building Footprints dataset \cite{MicrosoftBuildingFootprints}. 

\subsubsection{Preprocessing}
Prior to annotation, orthomosaics were tiled into one of two image sizes, 2048x2048 which would be sent to annotators or 8500x8500 which would be used for BDA bulk annotation and is described below. These two sizes were chosen because 2048x2048 contained what was believed to be a reasonable number of buildings per sample so as to not overload annotators, and 8500x8500 was only slightly less than the maximum image size permitted on the LabelBox platform. Tiling began in the top left of each orthomosaic and proceeded in steps of (tile size in pixels / 1.05) to create a slight overlap between tiles to ensure all pixels were annotated. In total 47 orthomosaics were tiled into 2048x2048 image tiles, 4 orthomosaics were tiled into 8500x8500 image tiles, and 1 orthomosaic (20210703-Champlain-Towers -South.geo.tif) was not tiled due there being no building polygons to annotate. Once the orthomosaics were tiled, building polygons, sourced from Microsoft Building Footprints \cite{MicrosoftBuildingFootprints}, were overlaid. These tiles and building polygons were uploaded to LabelBox\cite{Labelbox2024} for annotation.

\subsubsection{Annotation}
\label{subsec:annotation}
% BDA annotations consisted of annotating 2048x2048 image tiles within LabelBox. Annotators were intructed to label each builing polygon within the image tile based on the Joint Damage Scale (JDS) annotation schema, which unties builidng damage assessment scales across disaster types. There are five levels of damage classification, "no damage", "minor damage", "major damage", "destroyed", and "un-classified", and their explanations are provided in \cite{gupta2019xbd}. This process yeilded a total of 18,700 images annotated by a subset of 55 annotators from a greater annotation effort consisting of 130 annotators. 
% PRIYA: Integrated the above paragraph throughout this section

The annotation steps consisted of two initial annotations, BDA annotation, and BDA bulk annotation, done prior to the review steps, and two conditional annotations, spot check annotation and manual annotation, done post review steps. This section will discuss the two initial annotations and a discussion of the two conditional annotations will be presented in section \ref{subsec:review}.

The BDA annotations were generated by annotating 2048x2048 image tiles using LabelBox. Annotators were instructed to label each building polygon within the image tile based on the JDS annotation schema. %Joint Damage Scale (JDS) \cite{gupta2019xbd} annotation schema, consisting of five labels of damage, "no damage", "minor damage", "major damage", "destroyed", and "un-classified". 
This annotation effort yielded a total of 18,780 images annotated by a subset of 55 annotators from a greater annotation effort consisting of 130 annotators. 

% There were four types of annotations employed during the joint damage scale label generation: BDA annotations, BDA bulk annotations, spot check annotations, and manual annotations. Image tiles were either subject to BDA annotations or BDA bulk annotations, followed by spot check annotations and manual annotations, discussed in subsection \ref{subsec:review}. 

% The annotations utilized the JDS annotation schema, which unites building damage assessment scales across disaster types \cite{gupta2019xbd}. There are five levels of damage classification, ``no damage", ``minor damage", ``major damage", ``destroyed", and ``un-classified", and their explanations are provided in \cite{gupta2019xbd}. 

% The BDA annotations consisted of annotating 2048x2048 image tiles within LabelBox. Annotators were instructed to label each building polygon within the image tile based on the JDS annotation schema. Within this, a total of 18,780 image tiles were annotated by a subset of 55 annotators from a greater annotation effort consisting of 130 annotators. 

 BDA bulk annotations were performed for orthomosaics which were deemed by the authors to contain quantities of damage low enough that it would be inefficient to present them to annotators. Instead, all building polygons from relevant orthomosaics would be initially labeled as ``no-damage" and only building polygons that did not belong to that class would be manually labeled. This was done in an effort to not waste the time of annotators, and so three of the authors solely participated in BDA bulk annotation. This process exclusively used the 8500x8500 image tiles. This annotation effort yielded annotations for 177 image tiles. 

%For the purpose of ``bulk" annotating these image tiles, building polygons were provided with a default label of ``no damage", and annotators only annotated the building polygons that were not a ``no damage" label. 

%Once image tiles were annotated by one of the previously discussed annotation types, the annotations were combined at the orthomosaic level and subject to spot-check annotations made by a committee of reviewers during the review stages. The spot check annotations consisted of changing an previously annotated label to a label that is agreeable with the annotation schema.

%Lastly, the reconstructed orthomosaics with annotations were subject to potential manual annotations made by the committee of reviewers. The manual annotations consisted to drawing in additional building polygons and labeling the newly drawn building polygons. 

\subsubsection{Post Processing}
\label{subsec:postprocess}
After the BDA annotations and BDA bulk annotations, discussed in section \ref{subsec:annotation}, and an initial review, discussed later in section \ref{subsec:review}, the image tiles and their annotations were merged into orthomosaics, through a post process consisting of reconstructing orthomosaics with annotations. The reconstruction of orthomosaics with annotations consisted of mapping the annotations made on the image tiles to the original orthomosaic imagery. In cases where there were multiple annotations for the same building polygon, due to building polygons spanning multiple tiles,  annotations were merged by taking the ``highest" JDS damage label. 

\subsubsection{Review}
\label{subsec:review}
The review step consisted of a two-stage review process an attempt to reduce label noise and to ensure the correctness of annotations according to the JDS annotation schema. The two-stage review process consisted of an initial review of the image tiles, followed by a review of the reconstructed orthomosaics with annotations. This section further details these two stages of review in the order presented. 

% The correctness of the labels generated by the human annotators was paramount to this effort. In an effort to minimize label noise, and to ensure the correctness of annotations, a two-stage review process was employed. A visualization of this review workflow can be found in Figure \ref{Fig:workflow}.

In the first stage, the individual labeled image tiles were reviewed by one of the authors, and were verified for correctness. In the event that the annotations were incorrect, the labels were either corrected by the reviewer or requeued for annotation by a different annotator. This process continued for all annotations until all had been approved. 1,951 tiles (10.4\% of all tiles) were either corrected by a reviewer, or requeued.

In the second stage, and following the initial image tile review and the reconstruction of orthomosaics with their annotations, the annotations were reviewed at the orthomosaic level by a committee of reviewers. Corrections were made to any buildings which had been labeled incorrectly and in some cases, at the discretion of the committee, additional building polygons were manually added. 
%The postprocessing step, described in section \ref{subsec:postprocess}, presents an opportunity for label noise to enter as some building polygons spanned multiple tiles, the annotations of which needed to be merged as well. To mitigate this specific type of label noise, but also in hopes of maximizing label correctness overall, each orthomosaic was then reviewed and spot check annotation were conducted by 
The committee of reviewers, consisting of at least two members from a group of three authors and two external reviewers, corrected the labels of 1,285 building polygons (5.9\% of initially annotated building polygons).
Additionally, at this stage, the committee of reviewers also had the option to manually create and label new building polygons independent from those sourced from Microsoft Building Footprints \cite{MicrosoftBuildingFootprints}. This was done on a case-by-case basis for building types and building labels that were believed to be underrepresented in the existing data. 108 building polygons (0.5\% of all building polygons) were added through manual annotations.

% for these tiles were merged at the orthomosaic level, following the post process described in section \ref{subsec:postprocess}, and reviewed. The post process represents an opportunity for label noise to enter as some building polygons spanned multiple tiles, the annotations of which needed to be merged as well. 
% In an effort to mitigate this specific type of label noise, but also in hopes of maximizing label correctness overall, each orthomosaic was then reviewed by a committee of reviewers, consisting of least two authors.
% 1,285 building polygons (5.9\% of initially annotated building polygons) had their annotations changed during this second stage of review.
% Additionally, at this stage, the committee of reviewers also had the option to manually create and label new building polygons independent from the Microsoft Building Footprints. This was done on a case-by-case basis for building types and building labels that were believed to be underrepresented in the existing data. 104 building polygons (0.5\% of all building polygons) were added in this manner.

\begin{figure}
\centering
\includegraphics[width=0.66\columnwidth]{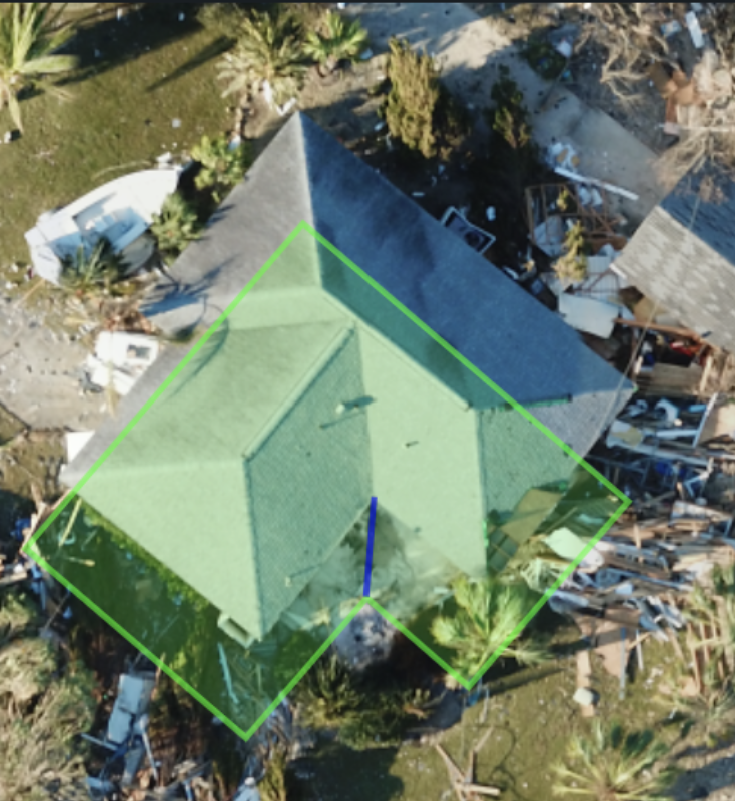}
\caption{The adjustment annotation for a building polygons to correct the alignment error, shown in \cite{manzini2024non}. The green polygon represents the building polygon sourced from Microsoft Building Footprints \cite{MicrosoftBuildingFootprints}, and the blue line represents the adjustment annotation made to correct the alignment error by connecting the current building polygon vertex to the correct location.}
\label{Fig:adjustment_introduction}
\end{figure}

\subsection{Spatial Alignment}
\label{sec:spatial_alignment}

Spatial alignment between the building polygons and the geospatial imagery within this dataset is an important component because even slight perturbations in spatial alignment can substantially impact ML model performance \cite{maiti2022effect, vargas2019correcting} for tasks like building damage assessment. As a result, controlling these alignment errors will enable more performant ML models to be trained, thereby enhancing the capabilities of ML models trained on this dataset. An example of the adjustment annotation to correct such alignment errors is shown in Figure \ref{Fig:adjustment_introduction}. This section discusses the sources of spatial misalignment, its importance to this dataset, and the spatial alignment correction annotations within the dataset.

\subsubsection{Necessity of Spatial Alignment}

The necessity of spatial alignment within this dataset is driven by three reasons: the presence of spatial alignment errors during the creation of this dataset, the dataset's motivation and downstream tasks, and to fill the gap within existing literature. This section further discusses these reasons in that order.

Spatial misalignment between the raw building polygons and the geospatial imagery were observed during the creation of this dataset. Spatial misalignment between the building polygons and the geospatial imagery can derive from five primary sources: satellite imagery acquisition, building polygon generation from satellite imagery, GSD variation between satellite imagery and sUAS imagery, sUAS GPS noise, errors with the raw imagery to orthomosaics generation process \cite{manzini2024non}. All of these sources were present within the creation of this dataset's imagery and BDA annotations, resulting in the problem and need to address spatial misalignment.

The correction of this spatial misalignment is relevant to this dataset's motivations and potential downstream tasks. As presented earlier, the creation of this dataset is motivated by the potential for it be used for disaster damage assessment by the ML and CV communities and its use within real-world disaster response. ML and CV efforts can be hindered by the presence of spatial alignment errors with a reduced performance deriving from spatial alignment errors \cite{manzini2024non}. Along with a reduction in performance for ML and CV efforts, spatial alignment cannot be left unaddressed; if there is the intention of integrating these efforts of disaster damage assessment in real-world disaster response, then it must be able to handle spatial alignment errors, due its inevitable presence in real-world scenarios \cite{manzini2024non}. 

\begin{figure}
\includegraphics[width=\columnwidth]{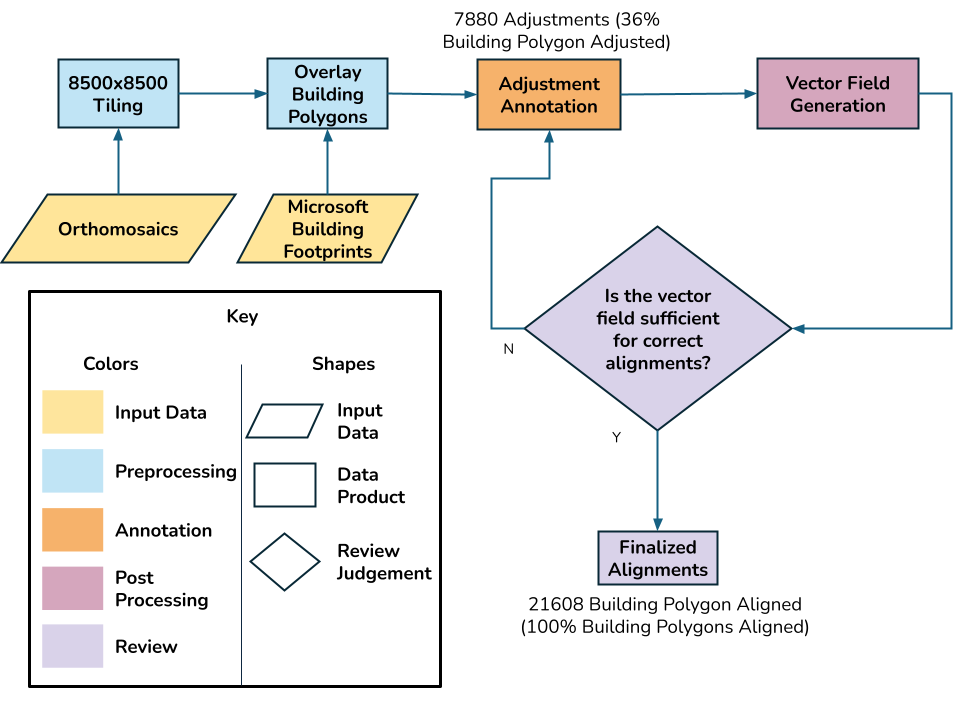}
\caption{The adjustment annotation and alignment vector field generation process to correct spatial alignment errors presented in \cite{manzini2024non}. For each orthomosaic, building polygon's were aligned with the vector fields generated through the curation of adjustment annotations that were further refined through reviews.}
\label{Fig:spatial_workflow}
\end{figure}

\begin{figure} [!h]
\centering
\includegraphics[width=\columnwidth]{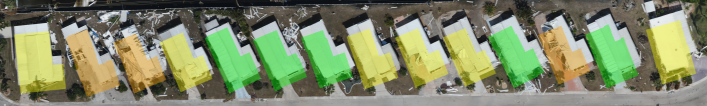} \\
\vspace{0.05cm}
\includegraphics[width=\columnwidth]{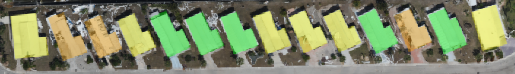}
\caption{An example of spatial alignment error correction with the vector fields generated through the process described in \cite{manzini2024non}. [Top] The raw, unaligned, building polygons (colored in green, yellow, and orange) sourced from Microsoft Building Footprints \cite{MicrosoftBuildingFootprints} overlaid over the geospatial imagery. [Bottom] The aligned building polygons (colored in green, yellow, and orange) derived from the generated vector fields from adjustment annotations. Note the difference in overlap in building polygons and the buildings within the imagery. }
\label{Fig:adjustment_unadjust}
\end{figure}

As presented in section \ref{sec:motivation_prev_work}, there is a gap within the literature on addressing the non-uniformity of spatial alignment errors, and there is no dataset that provides building spatial alignment corrections to be utilized to address such an issue. Therefore, any observation of spatial alignment error within this dataset would be hindered by the lack of previous efforts or lack of standard to follow. 

% However, the correction of spatial alignment is further hindered by the lack of existing literature and data sources for correcting such misalignment \cite{manzini2024non}. 

\subsubsection{Spatial Alignment Error Correction}

% As discussed earlier, the building polygons used in this work were sourced from the Microsoft Building Footprint Dataset\cite{MicrosoftBuildingFootprints} which themselves were sourced from satellite imagery and the orthomosaics were generated from sUAS raw imagery, leading to the misalignment discussed in \cite{manzini2024non}. This spatial misalignment can derived from sou

% Whereas, the imagery in this work was sourced from sUAS as described above, as a result of factors discussed in \cite{manzini2024non}, these building polygons do not consistently align with imagery.

%TODO: Mention that we are only dealing with translational errors here instead of rotational/scale errors. -- Added a sentence in the below paragraph. 

In order to correct these misalignments and to fill the gap within the literature, the building polygons within this dataset were manually aligned following the process described within \cite{manzini2024non}, a visual of this process is shown in Figure \ref{Fig:spatial_workflow} and result of this process is shown in Figure \ref{Fig:adjustment_unadjust}, consisting of five steps: input data, prepossessing, annotation, post processing, and review. The input data used for this process consisted of 51 orthomosaics, this excludes the ``20210703-Champlain-Towers -South.geo.tif" orthomosaic as this orthomosaic's building polygons were not derived from the Microsoft Building Footprints \cite{MicrosoftBuildingFootprints} and therefore not subject to the same spatial alignment errors as the other orthomosaics, and the building polygons from Microsoft Building Footprints \cite{MicrosoftBuildingFootprints}. Next, all 51 orthomosaics were preprocessed through tiling them into 8500x8500 image tiles and then overlaid with the building polygons from Microsoft Building Footprints \cite{MicrosoftBuildingFootprints}. After the preprocessing steps, adjustment annotations were provided for the building polygons to correct any spatial alignment errors present. It is worth noting that the adjustment annotations were only made to correct translational spatial alignment error, a reasoning for this is provided in section \ref{sec:limitations}. This step resulted in 7,880 adjustment annotations, corresponding to 36\%  of building polygons adjusted. These adjustments represent pixel-based offsets that are used to shift the polygons into the correct pixel coordinates in this imagery. More specifically, these adjustment labels are used in the post processing step to populate a vector field, which can be used to align all building polygons in an orthomosaic. Each of these vector fields were reviewed for their ability to align the building polygons within the orthomosaics, with more adjustment annotations added until the vector field was sufficient to align the building polygons. After this review step, all 21,608 building polygons sourced from Microsoft Building Footprints were aligned.

% As a result of this process, total of 7,880 adjustment annotation were collected for aligning the building polygons to the imagery within this dataset. 
% In order to correct for these misalignments, adjustment labels were generated by the authors to bring the building polygons into alignment with the imagery. This collection effort resulted in 7,880 adjustment annotations. 
% These adjustments represent pixel-based offsets that are used to shift the polygons into the correct pixel coordinates in this imagery. More specifically, these adjustment labels are used to populate a vector field, which can be used to align all building polygons in an orthomosaic. The adjustment annotation and alignment processes are further described in detail in \cite{manzini2024non}.

\subsection{Dataset Composition}
\label{sec:dataset_details}

The statistical distributions of the events,  labels, and errors for the dataset plus the rationale used for generating the train and test split merit further discussion. The dataset deliberately does not provide a validation split. 

%Since understanding the distributions present are critical for effective modeling to take place, this portion of the document will describe the contents of the dataset \ref{sec:dataset_stats} and its division into training and testing sets \ref{sec:train_test}. 

%One is to enable rigorous and structured evaluation of models derived from the dataset. As discussed in
%Sec.~\ref{?}, the CRASAR-U-DRIODs dataset is divided into a train (6 disasters) and test (4 disasters) split, and consists of varying? IMPRECISE distributions of classes. The second is to confirm that the dataset is WHAT? Labels/classes are balanced?

% NOT HELPFUL ENOUGH This portion of the document details the dataset's train and test split, as well as the distributions of buildings, area in pixels and square miles, and labels in the combined dataset.
\subsubsection{Dataset Statistics}
\label{sec:dataset_stats}

%\textbf{WHAT IS TAKEAWAY?} - Tom: I see the point here as describing the distributions that potential users of the dataset would encounter when using it. Since this is a dataset paper I don't think we are trying to prove anything here, or argue for some specific claims, we are just trying to say what is in the dataset at a high level.

The distribution of buildings, events, labels, and adjustments present in this dataset will be discussed below in order to detail the distribution that this dataset represents. 

\begin{figure} 
\includegraphics[width=0.5\textwidth]{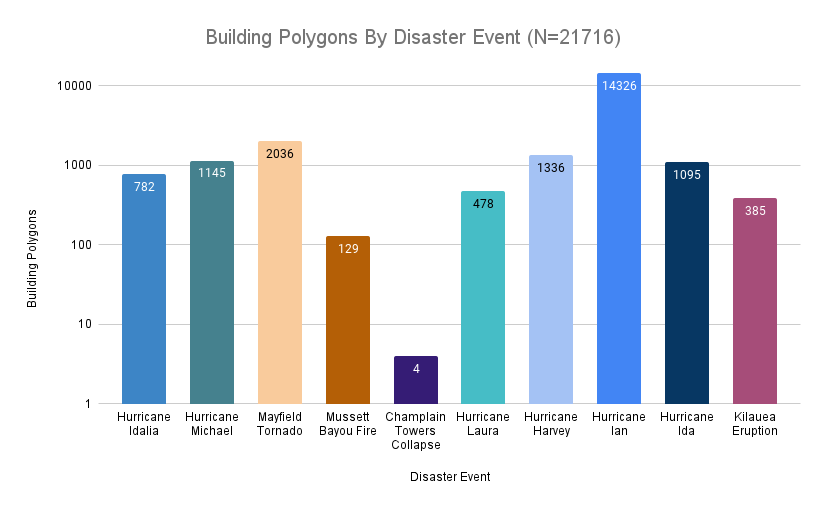}
\caption{Distribution of building polygons across the ten federally declared disaster events. Note y-axis log scale.}
\label{Fig:building_events}
\end{figure}

Each of the ten disasters within the dataset contains different counts of building polygons; a visual of the building polygon distribution across the disaster events is shown in Figure \ref{Fig:building_events}. Hurricane Ian has the highest number of building polygons with 14,326 presented within this disaster event, and the Champlain Towers Collapse has the lowest number of building polygons with 4 building polygons. %A visual of the building polygon distribution across the disaster events in shown in Figure \ref{Fig:building_events}.

Hurricane Ian is represented the most within the dataset in terms of pixel count, with 30.74 gigapixels. The remaining nine disaster events make up the remaining 36.295 gigapixels within the dataset. Similarly, Hurricane Ian is represented the most within the dataset in terms of area with 32.67km\textsuperscript{2}, compared to the remaining nine disaster events that make up at combined total of 35.31km\textsuperscript{2}. 

\begin{figure} 
\includegraphics[width=0.5\textwidth]{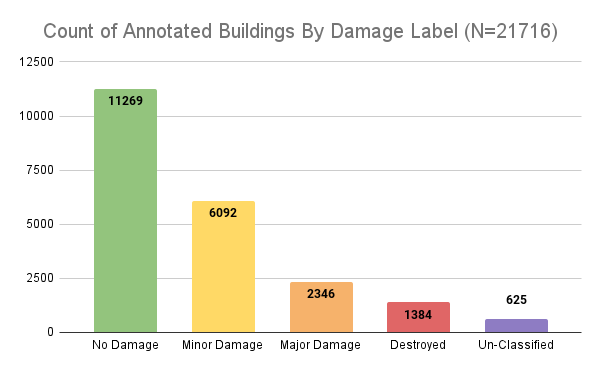}
\caption{Distribution of building polygon damage labels across the five JDS damage classes (No Damage, Minor Damage, Major Damage, Destroyed, Un-Classified).}
\label{Fig:datastat_buildings}
\end{figure}

The dataset's majority class is ``no damage"  with 11,269 buildings, and the most underrepresented damage label is ``un-classified", with 625 buildings. The remaining damage labels, ``minor damage", ``major damage", and ``destroyed", are 6,092 buildings, 2,346 buildings, and 1,384 buildings, respectively. Figure \ref{Fig:datastat_buildings} shows the class distribution for building damage labels.

In addition to the building damage assessment labels, the dataset's adjustment annotations for spatial alignment present a non-uniform pattern \cite{manzini2024non}. As discussed in \cite{manzini2024non}, these spatial alignment errors vary on an orthomosaic and disaster event level, with no prominent observation of a normal distribution in the spatial alignment errors that occur. 

\subsubsection{Train \& Test Split}
\label{sec:train_test}

\begin{figure}[]
\centering
\includegraphics[width=0.98\columnwidth]{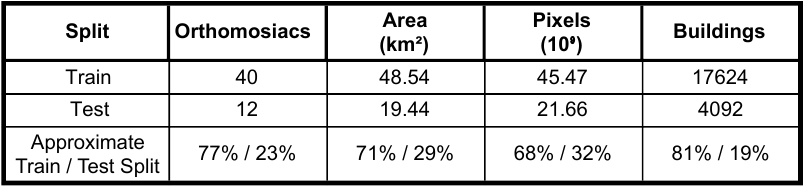}
\captionof{table}{Comparisons between the train and test split in terms of count of orthomosaics, square kilometers, gigapixels, and count of building polygons.\label{tab:traintestsplit}}
\end{figure}

%WHY INDEPENDENT OF DISASTERS?

%  arg 1: temporally; train on past ex 2018, then test after it.  WHY NOT-- later Ian and Idalia have huge datasets, bigger disaster and bigger use of drones not even distribution of drones or proportional, it would only be hurricanes.

%  arg 2: geographical split, ex. east, west  don't have the data since hurricanes tend to hit florida

%  arg 3: disaster type split: ex. hurricanes, then test on everything else. sufficient coverage for both test and training (shortcoming, so leaves out earthquakes and tsunamis)

%  subjective: Tried to balance disaster types, image quantity. Could be done differently
  
Table \ref{tab:traintestsplit} summarizes the train and test splits for the dataset. The split chosen here is made at the disaster level, meaning that all orthomosiacs from a disaster are contained in either the test or train set. The train set consists of all orthomosaics collected at Hurricanes Harvey, Ian, Laura, and Ida, the Kilauea Volcano Eruption, and the Champlain Towers Collapse. The test set consists of all orthomosaics collected at Hurricanes Idalia and Michael, the Mussett Bayou Fire, and the Mayfield Tornado. Ideally, a train and test split would mirror the operational use case that any trained ML models would experience in practice. With this in mind, the split represents a subjective choice balancing the disaster type and quantity of orthomosaics given the available data. While there are strict alternative strategies such as temporal sampling (eg. train on the past, test on future), spatial sampling (eg. train on the east, test on the west), or uniform random sampling of orthomosaics, each choice permits data leakage between the test and train set. The decision quickly becomes about which leaks are tolerable and what train/test balance does the strategy create.

% test set do not share specific events but do share disaster types (2 hurricanes)
Although the train and test sets chosen do not overlap with respect to specific disasters, they do overlap temporally. For example, in the test set, Hurricane Michael occurred in 2018, while in the train set, Hurricane Ian occurred in 2022. It appears that there is no way to organize this data such that a test set remains both temporally and disaster-independent of the train set while also being valid for evaluations of trained ML systems. As a result, the train and test split presented here represents a reasonable compromise while remaining independent of disasters. %The complete details of this split are shown in Table \ref{tab:traintestsplit}.

This dataset intentionally does not provide a validation split. Model validation for systems trained on this dataset represents an area of exploration, and publishing a validation set may constrain those who wish to validate their models in different ways. As a result, model validation is left as an exercise for the reader.

%further analysis of the spatial alignment data is discussed in \cite{manzini2024non}.

\section{Limitations}
%TODO: Expand and cleanup
\label{sec:limitations}
There are four limitations in the CRASAR-U-DRIODs dataset: the imagery is limited to the United States, not all disaster types are represented, damage assessment from aerial imagery does not necessarily represent ground truth building damage labels, and rotational, scale, shape based alignment errors are not considered. Although these are shortcomings, it should be noted that the first three stem from the availability of source imagery with provenance and operational fidelity. The fourth, that only translation spatial alignment errors were corrected, was a pragmatic decision on how to best allocate personnel time given that the non-translational alignment errors did not appear to be as severe. 

The imagery within this dataset is limited to disaster-affected areas within the United States, presenting concerns of generalization of downstream models to other geographical regions. Although this geographical limitation may prevent expanding efforts to generalize to a greater variation of geographic locations, the dataset provides more coverage in terms of area, providing more variation in geographic location compared to all other disaster sUAS disaster datasets, which are also limited to individual countries, as discussed in section \ref{sec:aerial_imagery_disaster_response}. 

The ten federally declared disasters represented within this dataset are limited to five disaster types, lacking representation of other disaster types, presenting another issue of generalization. 
%but, in comparison to prior sUAS disaster datasets, the dataset provides more variation among disaster types. 
As shown in Table \ref{Tab:dataset_split_by_disaster}, this dataset provides imagery from five disaster types: hurricane, tornado, volcano, wildland fire, and man-made collapse, which does not include other disaster types like earthquakes and tsunamis represented elsewhere in the literature. Although this is a limitation to consider when expanding any efforts with this dataset to other disaster types, this dataset provides the most variation in disaster types among sUAS disaster datasets, as discussed in section \ref{sec:aerial_imagery_disaster_response} and detailed in Table \ref{Tab:dataset_disaster_types}. 

The building damage assessment labels within this dataset are based on viewpoints from aerial imagery and do not necessarily reflect the actual conditions of the building. This presents a potential inconsistency in the ground truth building damage assessment labels and the aerial-based building damage assessment labels. This is an inherent limitation with labels generated from aerial imagery and is a limitation of all datasets discussed in section \ref{sec:motivation_prev_work}. Further, it is unclear if training based on labels generated based on ground level or interior inspections of buildings would represent a reasonable target function for downstream ML models, as these labels would depend on information that would not be available to any ML models consuming aerial imagery at inference time.

%Within the dataset's imagery, the buildings structures may be obscured by debris, vegetation, tarps, or another part of the structure. In these cases, the building damage assessment class was inferred based on the conditions that may be present behind the obscuration. Inspection done on ground level or interior inspections would provide the ground truth damage class; however, this approach would require more manual effort and time to complete. Given, the downstream task of this dataset to be used for ML and CV efforts that will automate building damage assessment for disaster response, the reliance on aerial imagery aligns more closely with the operational use case. 

The building alignment provided by the adjustments within this dataset is limited to translational alignment, excluding rotation, scale, and shape-based errors. While this does not address all possible spatial alignment errors, it represents a starting point for addressing these other error types. 
%While alternatives exist, such as modifying the original imagery such that it aligns with the does align with downstream tasks. The reasoning behind this is for downstream tasks, like building damage assessment, the spatial map should be reconstructed to its original state. Therefore, modifying the orthomosaics through skewing or rotation does not contribute to these downstream tasks.

% First, this analysis was limited to the translational errors, meaning measurements of skew or rotational errors were not included. This study represents an initial attempt to quantify these alignment errors, and these additional error types should be explored in future work.
% The reasoning behind this is for downstream tasks, like building damage assessment, the spatial map should be reconstructed to its original state. Therefore, modifying the orthomosaics through skewing or rotation does not contribute to these downstream tasks.

% United states only

% Disasters that we have only (no earthquake, no tsunami, etc)

% BDA labels are based on the imagery not a ground level/interior inspection of the building.

\section{Conclusions}
\label{sec:contributions}

The CRASAR-U-DROIDs dataset represents the largest collection, in terms of pixels, of labeled orthomosaics collected from sUAS known at this time, not just disasters. The dataset is particularly valuable for disaster research and application development because the source imagery is drone flights tasked by agencies having jurisdiction for each disaster, thereby providing operational fidelity. All imagery within the dataset has been screened to exclude human remains,  private personal information, or any other content not approved by the agencies for general release to the public. This dataset opens new opportunities for transparent and ethical ML model training for sUAS imagery at a scale that has not been explored before.

%The release of the CRASAR-U-DROIDs dataset represents the largest collection of labeled orthomosaics collected from sUAS known at this time, not just disasters. This dataset was sourced from data collected within disaster areas as part of response and recovery activities  at the direction of agencies having jurisdiction. This dataset opens new opportunities for ML model training for sUAS imagery at a scale that has not been explored before. The CRASAR-U-DRIODs dataset is publicly available at \url{https://huggingface.co/datasets/CRASAR/CRASAR-U-DROIDs}, and the raw data used to generate the orthomosaics within this dataset can be obtained by contacting the authors. All imagery within the dataset has been screened to not include human remains or any other explicit content, and the provenance of the imagery is known. 

Independently of the dataset, this article provides disaster-oriented and sUAS researchers with a survey and analysis of existing computer vision/machine learning datasets. It also describes a
design pattern for collecting, annotating, and releasing sUAS datasets with clear provenance and transparency. 

CRASAR-U-DROIDs specifically contributes to the larger machine learning, computer vision, and remote sensing communities as well as the robotics and the emergency management communities. In reverse order, it supplies
the emergency management community with  images can be used to train emergency managers and sUAS pilots as to what to collect, and lables which give examples of the types of outputs that downstream machine learning systems might produce.
The dataset is expected to especially benefit the robotics and machine learning, computer vision, remote sensing, and emergency management communities as follows:
\begin{itemize}
  \item The largest dataset of disaster imagery from sUAS which will enable the development of machine learning models for building damage assessment to the benefit of the emergency management community.
  \item The largest dataset of sUAS orthomosaic imagery in terms of pixels, and buildings to the benefit of the ML and CV communities.
  \item The first dataset of geospatial imagery that explicitly controls for non-uniform spatial alignment errors between the imagery and geospatial data to the benefit of the robotics and ML, CV, and remote sensing communities.
  \item The first effort to bridge the gap between satellite and sUAS spatial imagery for machine learning systems by utilizing the same classification schema as relevant prior work \cite{gupta2019xbd} to the benefit of the ML and CV communities.
  \item The application of the Joint Damage Scale (JDS) for drones, which enables future transfer and comparison with satellite models. 
\end{itemize}

It is hoped that the ultimate contribution of CRASAR-U-DROIDs will serve as the basis for models that will revolutionize disaster response. Ongoing and future work associated towards that goal is focusing on three topics:  the development of machine learning models that can jointly perform alignment (including  rotation, scale, and shape-based errors) and damage assessment for buildings, the labeling of roads and their levels of passability and obstruction based on this aerial imagery, and the annotation of coincident imagery of these same scenes taken from satellite and manned aircraft to enable multiview and multiscale ML models to be trained.

The CRASAR-U-DRIODs dataset is publicly available at \url{https://huggingface.co/datasets/CRASAR/CRASAR-U-DROIDs}, and additional labels will be added there as ongoing research progresses. The raw data used to generate the orthomosaics within this dataset can be obtained by contacting the authors. 

%Current efforts are improving this dataset and involve collecting and curating labels for road conditions. Additionally, parallel data from manned and satellite imagery of the same scenes is being annotated. Regarding ML modeling, future work will focus on the development of performant ML models for building polygon alignment and building damage assessment. 

\section*{Acknowledgment}
This material is based upon work supported by the National Science Foundation under AI Institute for Societal Decision Making (AI-SDM), Award No. 2229881 and under ``Datasets for Uncrewed Aerial System (UAS) and Remote Responder Performance from Hurricane Ian," Award No. 2306453. The authors thank CRASAR and David Merrick for the acquisition of the imagery and supporting information, the 130 annotators and the instructors from the Winchester Thurston School, Ball High School, Bryan Collegiate High School, and Rudder High School for their annotation efforts, and Mihir Godbole and Hasnat Abdullah for their helpful reviews. 

\bibliographystyle{IEEEtranS}
\bibliography{references}

\begin{figure}[h!tb]

\centering
\includegraphics[width=\textwidth]{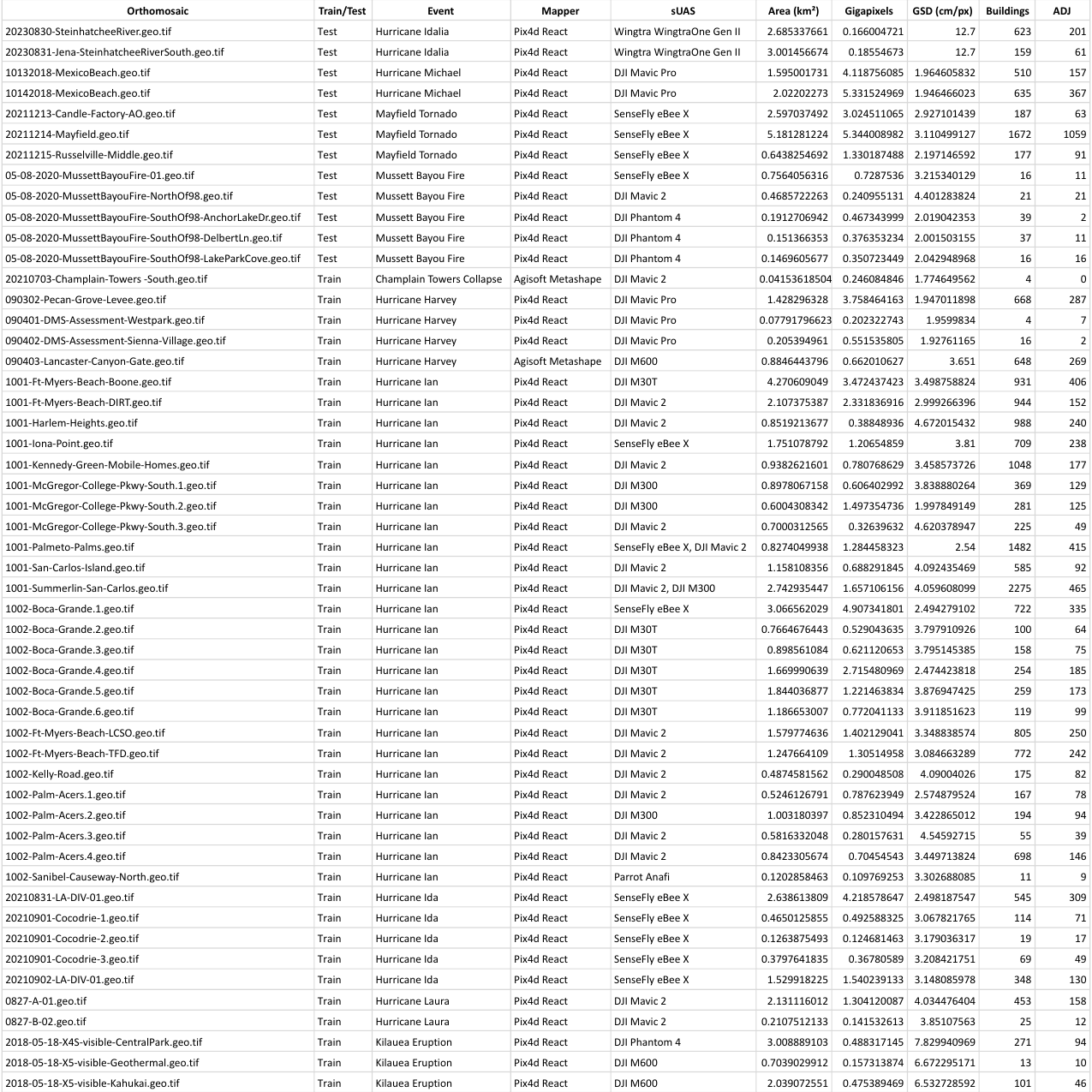}
\captionof{table}{CRASAR-U-DROIDs Overview. All 52 orthomosaics ordered by their split and disaster event. Information regarding the georectification software, the sUAS which captured the imagery, the area associated with the orthosmosiac, the number of pixels, the ground sampling distance, and the count of buildings and adjustments are included as well.}
\label{Tab:Dataset overview}
\end{figure}

\newpage
\onecolumn
\appendix
\centering
Table \ref{Tab:Dataset overview} contains the details of the orthomosaics in the dataset and their respective annotations.

\end{document}